\documentclass[11pt]{article}

\usepackage[final]{acl}

\usepackage{times}
\usepackage{latexsym}

\usepackage[T1]{fontenc}
\usepackage[utf8]{inputenc}

\usepackage{microtype}

\usepackage{inconsolata}

\usepackage{graphicx}

\usepackage{tikz}
\newcommand{\blackcircle}[1]{%
    \tikz[baseline=(char.base)]{%
        \node[shape=circle, fill=black, inner sep=0.5pt, text=white] (char) {#1};%
    }%
}

\usepackage{tabularx}
\usepackage{xcolor}
\usepackage{subcaption}
\usepackage{amsmath}
\usepackage{paralist}
\usepackage{makecell}
\usepackage{array}
\usepackage{multirow}
\usepackage{amssymb}

\usepackage{algorithm}
\usepackage{algorithmic}
\usepackage{booktabs}
\newcolumntype{L}[1]{>{\raggedright\arraybackslash}p{#1}}

\usepackage{newfloat}
\usepackage{listings}
\DeclareCaptionStyle{ruled}{labelfont=normalfont,labelsep=colon,strut=off} % DO NOT CHANGE THIS
\floatstyle{ruled}
\newfloat{listing}{tb}{lst}{}
\floatname{listing}{Listing}

\title{PuzzleClone: A DSL-Powered Framework for Synthesizing Verifiable Data}

\author{Kai Xiong\thanks{Equal contribution}\thanks{Corresponding author} \\
  HiThink Research \\
  \texttt{xiongkai@myhexin.com} \\\And
  Yanwei Huang\footnotemark[1] \\
  HKUST \\
\texttt{yhuangix@connect.ust.hk} \\\And
  Rongjunchen Zhang\footnotemark[2] \\
  HiThink Research \\\quad
  \texttt{zhangrongjunchen@myhexin.com} \\
  \AND
  Kun Chen \\
  HiThink Research \\
  \texttt{chenkun2@myhexin.com} \\\And
  Haipang Wu \\
  HiThink Research \\
  \texttt{wuhaipang@myhexin.com} \\\And
  Yingcai Wu \\
  Zhejiang University \\
  \texttt{ycwu@zju.edu.cn} \\}

\begin{document}
\maketitle
\begin{abstract}
High-quality mathematical and logical datasets with verifiable answers are essential for strengthening the reasoning capabilities of large language models (LLMs). While recent data augmentation techniques have facilitated the creation of large-scale benchmarks, existing LLM-generated datasets often suffer from limited reliability, diversity, and scalability. To address these challenges, we introduce PuzzleClone, a formal framework for synthesizing verifiable data at scale using a novel DSL-driven approach.
Our approach features three key innovations: (1) encoding seed puzzles into structured logical specifications, (2) generating scalable variants through systematic variable and constraint randomization, and (3) ensuring validity via a reproduction mechanism. Applying PuzzleClone, we construct PC-83K, a benchmark comprising over 83K diverse and programmatically validated puzzles. The generated puzzles span a wide spectrum of difficulty and formats, posing significant challenges to current state-of-the-art models. Experimental results show that post training (SFT and RL) on PC-83K  yields substantial improvements not only on the testset but also on various logic and mathematical benchmarks. Post training raises average performance on PC-83K from 14.5 to 66.0 and delivers consistent improvements across 7 logic and mathematical benchmarks up to 18.4 absolute percentage points (SATBench from 51.6 to 70.0). 
Our code and data are available at \url{https://github.com/HiThink-Research/PuzzleClone}.
\end{abstract}

\section{Introduction}
Large-language models (LLMs) have recently demonstrated impressive zero-shot and few-shot performance on a wide spectrum of reasoning tasks, yet consistently achieving robust logical reasoning remains an open challenge. To push the frontier, researchers have released a variety of mathematical-puzzle and formal-logic benchmarks that expose models to carefully crafted, high-difficulty problems. Unfortunately, the manual effort required to compose and validate such items has kept existing corpora relatively small and homogeneous, impeding further progress.

One promising approach to address this bottleneck is \textit{data augmentation}, which involves generating new problems by systematically modifying existing ``seed'' instances~\cite{feng-etal-2021-survey, Lu2024}. This strategy has the potential to vastly expand the size and diversity of reasoning datasets while reducing the manual effort involved in their creation. However, current augmentation pipelines largely rely on LLMs to annotate new problems, generate solutions, and verify the answers~\cite{Lu2024, tan-etal-2024-large, Shah2024}, which introduces several critical limitations. First, without a robust, end-to-end verification approach, it is hard to ensure the data reliability throughout the synthesis pipeline, leading to flawed, inaccurate, or biased data~\cite{Wang2024}. Second, existing pipelines lack formalization and disproportionately depend on the generative capabilities of the underlying LLMs, which threatens data diversity. For example, an individual model typically explores only a narrow range of variations in assumptions, conditions, parameters, and queries. This limited coverage reduces the dataset’s ability to challenge and generalize LLM reasoning capabilities effectively. Finally, the substantial computational costs associated with involving LLMs in every step of the synthesis pipeline also severely constrain scalability~\cite{Wang2024}. Motivated by the need for a scalable path to unbounded yet trustworthy data creation, our work seeks a principled procedure that can, in principle, generate reliable reasoning data indefinitely.

\begin{figure}[t]
  \centering
  \includegraphics[width=\linewidth]{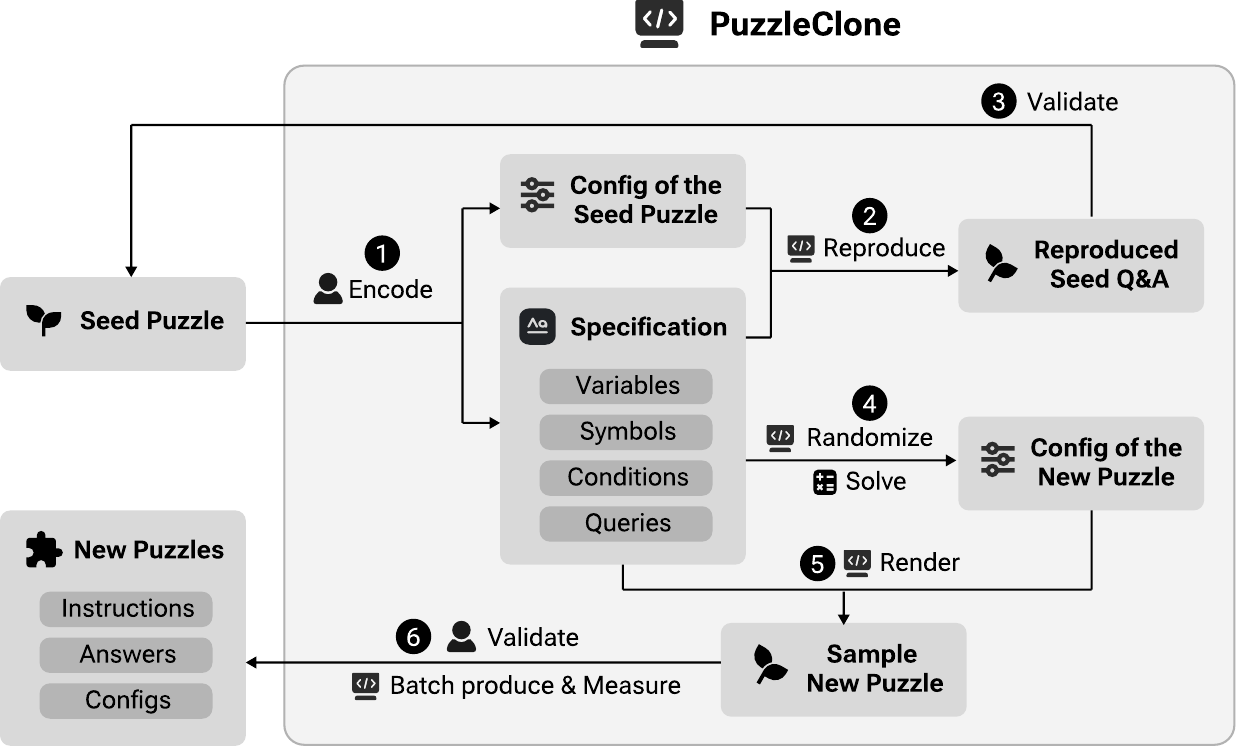}
  \caption{Overview of the PuzzleClone framework.}
  \label{fig:overview}
\end{figure}

% In this work, we introduce PC-83K, a novel benchmark of 83,657 challenging, diverse, and fully verifiable puzzles, covering  diverse puzzle types such as Satisfiability Modulo Theories (SMT), Linear Programming (LP), and classical logic puzzles like Sudoku. 
% The benchmark is generated via PuzzleClone, a new data curation framework featuring a novel Domain Specific Language (DSL) capable of modeling complex puzzle intricacies, alongside a rigorous pipeline to ensure programmatic validity of solutions. 

In this work, we introduce PuzzleClone, a new data curation framework featuring a novel Domain Specific Language (DSL) capable of modeling complex puzzle intricacies, alongside a rigorous pipeline to ensure programmatic validity of solutions. Utilizing this framework, we construct PC-83K, a novel benchmark of 83,657 challenging, diverse, and fully verifiable puzzles, covering diverse puzzle types such as Satisfiability Modulo Theories (SMT), Linear Programming (LP), and classical logic puzzles like Sudoku. 
Figure \ref{fig:overview} shows an overview of PuzzleClone. Each seed puzzle is first manually encoded into a structured problem specification that specifies the data synthesis pipeline, with a config file that contains the values of the parameters specific to the seed puzzle (\blackcircle{1}). Based on them, an instance generator can automatically batch produce new questions with diverse mathematical configurations by systematically varying parameters and combinations of constraints with randomization (\blackcircle{4}, \blackcircle{5}). Meanwhile, it can programmatically derive ground-truth answers for each generated instance with the assistance of symbolic or custom solvers (\blackcircle{4}).  To ensure fidelity, our pipeline includes a reproduction step that verifies the original seed questions can be regenerated from the DSL specifications (\blackcircle{2}, \blackcircle{3}), thereby guaranteeing the accuracy of both problem formulations and associated answers.

Our experiments demonstrate that PC-83K poses substantial challenges for state-of-the-art large language models (LLMs), including ChatGPT-4o and DeepSeek-R1. We further use it to distill the reasoning capabilities of large models into small ones. After post-training, Qwen2.5-7B-Instruct improves from 14.5 to 66.0 on PC-83K, and achieves gains of up to 18.4 absolute percentage points across 7 logic and mathematical benchmarks.

\section{PuzzleClone}
\label{sec:framework}

% \subsection{Motivating Cases}

% \subsection{Data Synthesis Pipeline}

\begin{figure*}[htbp]
  \centering
  % \scalebox{0.86}{
  \includegraphics[width=\linewidth]{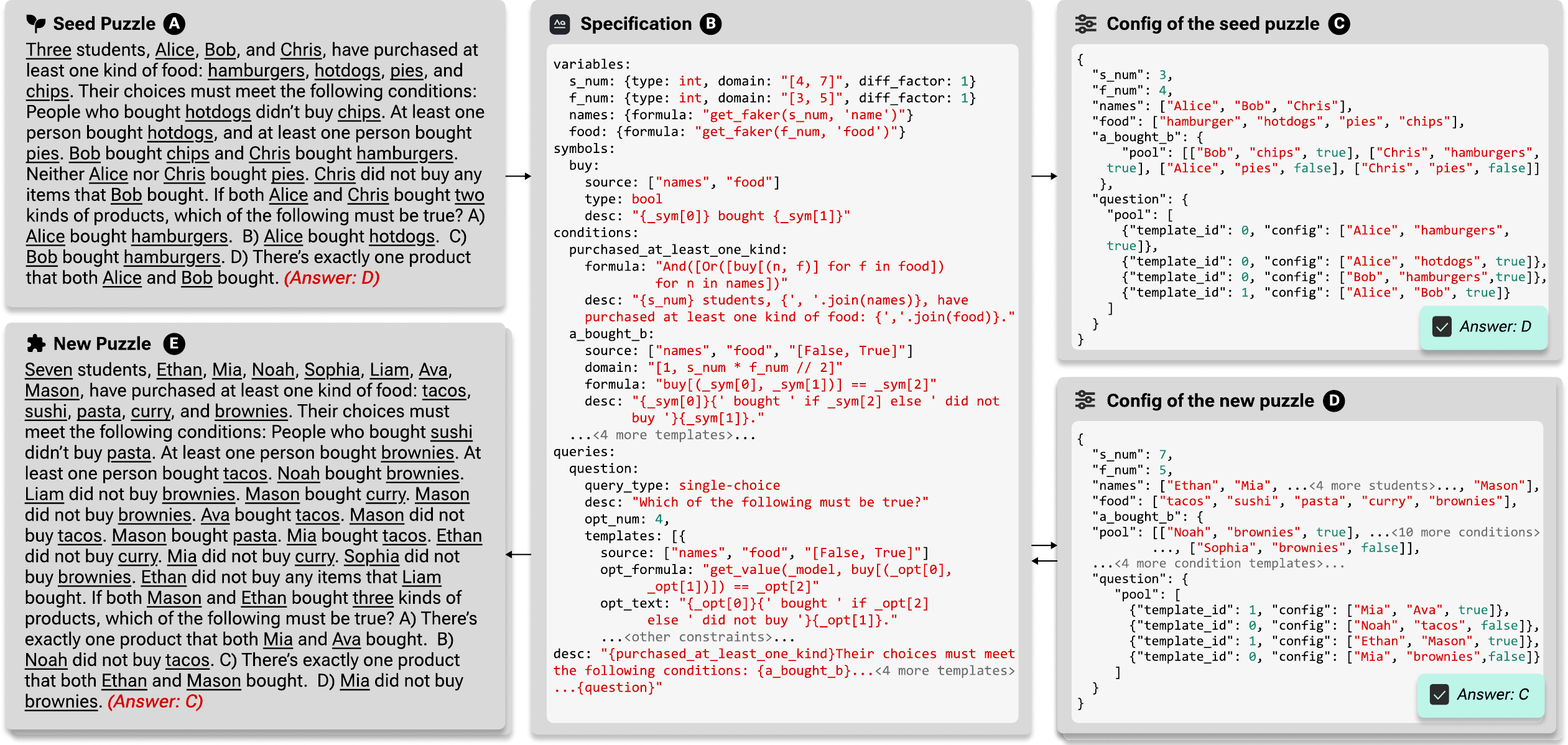}
  % \caption{The data synthesis pipeline of PuzzleClone. (A) The seed puzzle. (B, C) The DSL specification and the config files encoded from the seed puzzle. (D) An example config file of the new puzzles created from the puzzle generator. (E) An example of the new puzzle.}
  \caption{The data synthesis pipeline of PuzzleClone. (A) An example seed puzzle. (B, C) The DSL specification and the config encoded from the seed puzzle. (D) A randomly generated config produced by the puzzle generator for this DSL specification. (E) A new puzzle instance synthesized by the puzzle renderer using the DSL specification and the generated config.}
  \label{fig:pipeline}
\end{figure*}

The data synthesis process can be formalized as follows: For each input seed puzzle $P^0$, our goal is to generate a set of new puzzles $P^* = (I^*, A^*)$, where $I^*$ is the set of puzzle instructions and $A^*$ is the set of answers. The process comprises three stages: puzzle encoding, puzzle generation, and config-based validation.

\subsection{Puzzle Encoding}
Most puzzles can be conceptualized as a combination of a universal puzzle template, which encapsulates the core logic of the puzzle, and a set of parameters that define its specific details~\cite{pan-etal-2023-logic}. For example, in the seed puzzle illustrated in Figure \ref{fig:pipeline}A, the underscored texts represent the puzzle-specific parameters while the rest describe the template. Thus, data augmentation can be viewed as systematically varying these parameters and embedding them into the universal template to generate new puzzles.

During puzzle encoding, each seed puzzle $Q^0$ is manually encoded into a structured problem specification $Q_s$ and a configuration file $Q_c$. Contrary to prior works that rely on logic programming languages such as Prolog and SMT-LIB~\cite{clocksin2003programming, pan-etal-2023-logic, barrett2010smt}, we design a new domain-specific language (DSL) to describe $Q_s$, as shown in Figure \ref{fig:pipeline}B. This DSL not only captures the core puzzle logic but also encodes constraints that are implicit and only applicable to data augmentation, such as the parameters' value domains and the number of constraints. Furthermore, the DSL represents the puzzle in a more human-readable format while still machine-parsable for downstream processing. The DSL specification $Q_s = (V, S, C, Q, D)$ has the following components\footnote{See Appendix \ref{sec:schema} and \ref{sec:appendix_specs} for DSL schema and puzzle specs.}:

\noindent \textbf{Variables} $V$: Parameters that can be varied to generate new puzzles. For instance, variables in the seed puzzle in Figure \ref{fig:pipeline}A include the number of students \texttt{s\_num} and food types \texttt{f\_num}, as well as their names \texttt{names} and \texttt{food}. Each variable $v$ is characterized by its type (\texttt{type}) and value domain (\texttt{domain}), or a formula (\texttt{formula}) deriving it from other variables. For instance, \texttt{get\_faker} is an internal function which leverages Faker\footnote{https://github.com/joke2k/faker} to generate random natural-language-based instance names. Finally, each variable is assigned a \textit{difficulty factor} (\texttt{diff\_factor}) that specifies how the variable's value contributes to the puzzle difficulty ($+1$, $-1$, $0$ for positive, negative, and neutral effect.).
    
\noindent \textbf{Symbols} $S$: Quantities to be solved in the puzzle. Symbols of various types (\texttt{type}) can be created by mapping existing variables (\texttt{source}), and each symbol can be binded with a natural language (NL) template (\texttt{desc}) that outlines its expression in the puzzle instructions. For instance,  a set of boolean symbols \texttt{buy} are generated for each combination of students' and food names representing the purchase information, as depicted in Figure \ref{fig:pipeline}B.

\noindent \textbf{Conditions} $C$: A pool of all potential constraints that can be applied to the puzzle, including \textit{static conditions}, which remain the same among all puzzles, and \textit{dynamic conditions}, which are randomly generated from a template and can appear a variable number of times. For example, the statement ``\textit{[students]  have purchased at least one kind of food}'' in the seed puzzle in Figure \ref{fig:pipeline}A can be viewed as a static condition, while the subsequent text under ``\textit{their choices must meet the following conditions}'' outlines dynamic conditions with reusable structures and mutable parameters for replication.
    Our DSL provides structural declarations for randomly generating such conditions. 
    First, the dynamic parameters within each condition can be randomly determined according to the specified pool (\texttt{source}) and the allowable range for the number of instances of each condition type (\texttt{domain}).
    Additionally, every condition is associated with a formula template (\texttt{formula}) and a NL template (\texttt{desc}). The randomly chosen parameters are integrated into the templates to generate the complete condition.

\noindent \textbf{Queries} $Q$: The questions in the puzzle. For example, the \texttt{question} in Figure \ref{fig:pipeline}B presents a single-choice question (\texttt{query\_type}) with four options (\texttt{opt\_num}). The structure of these options is defined by two distinct templates (\texttt{templates}), each with its own parameter pool (\texttt{source}), formula and instruction template (\texttt{opt\_formula}, \texttt{opt\_text}).

\noindent \textbf{Description} $D$: An NL template assembling previous components for the eventual puzzle (\texttt{desc}).

In parallel with the specification, the puzzle-specific values of variables and the parameters within constraints and queries are extracted as a configuration file $Q_c$, as shown in Figure \ref{fig:pipeline}C. Thus, a puzzle can be uniquely determined by combining a specification and a configuration.

% $Q_s = (P, U, C, Z, Q)$ is written in a domain-specific language (DSL) designed by us, explaining the variable parameters $P$, the unknown quantities to be solved $U$, the pool of all potential constraints $C$, additional constraints on the solver $Z$, and the query $Q$ that describes the question to be answered. The configuration files $Q_c$ contain the values of the parameters, such as the size of the grid, the number of pre-filled cells, and so on.

\subsection{Puzzle Generation}
After encoding the seed puzzle, a puzzle generator will translate the specification into new puzzles. Specifically, it first generates a Python script which automatically decides all variables, conditions, and queries by random while leveraging either built-in symbolic solvers (e.g., Z3~\cite{z3} for SMT problems) or custom solvers to solve the puzzle. Thus, new puzzles can be generated by executing the script.

\subsection{Config-based Answer Validation}
One of the key goals of PuzzleClone is to keep each synthesis step verifiable for the reliability of the synthesized data. Since the puzzle encoding process relies on careful manual effort, it is necessary to make sure the process is accurate, leading to our design of a config-based answer validator. Specifically, a validation script is generated in parallel with the puzzle generator, which is capable of reproducing the seed puzzle by acquiring the puzzle-specific values directly from the config instead of randomization. Thus, a deterministic reference answer can be computed and compared against the ground-truth solution of the seed puzzle, thereby verifying the correctness of the encoding process.

% maybe we also need to compare the instructions to be more rigorous.

% \subsubsection{Data Polish by LLMs}

% \subsection{Grammar (DSL)}

\section{Dataset Construction}
% We constructed our dataset through a multi-stage process designed to ensure quality, diversity, and verifiability. 
We constructed our dataset via a multi-stage process to ensure quality, diversity, and verifiability.
% Below we detail the systematic process from raw seed collection to final dataset partitioning.

\subsection{Seed Puzzle Collection and Curation} \label{sec:seed_collection}
We started by collecting raw puzzles from diverse sources, including mathematics competitions for primary and secondary schools, Olympiad-style problems, logical reasoning books/blogs, and established puzzle benchmarks, e.g., Sudoku and combinatorial optimization problems~\cite{mittal_puzzlebench_2024, chen_lr2bench_2025, li2025logic, liu2025combibench}.
We then employed the Qwen2.5-72B-Instruct model with specially crafted prompts (see Appendix \ref{sec:prompts}) to select potential seed puzzles based on two key criteria:
% (1) The problem statement must contain variable values or constraints that can be randomly modified while maintaining the problem's validity; (2) The problem must be solvable by SMT within reasonable time and space complexity, with verifiable Python implementations.
\begin{compactitem}
    \item The problem contains variable values or logical constraints that can be replicated by random without altering its semantic validity.
    \item The puzzle is solvable using symbolic or custom solvers within tractable time and memory bounds, and a correct Python solution can be generated by the model.
\end{compactitem}

\subsection{Human Verification and Enhancement}
% The automatically filtered puzzles underwent thorough manual review by our research team. During this stage, we performed several quality improvement operations:
% \begin{itemize}
%     \item Conducted brainstorming sessions to enhance problem complexity through additional variables (e.g., A1-ant in our dataset) and novel constraints (e.g., A2-athlete)
%     \item Introduced diversified questioning approaches to increase variety
%     \item Eliminated problems with conceptually similar solutions to ensure mathematical diversity
% \end{itemize}
The AI-filtered puzzles underwent rigorous manual review, during which we (1) conducted brainstorming sessions to enhance problem complexity through additional variables and novel constraints (e.g., \texttt{A1-ant} vs. \texttt{A2-athlete} in our dataset), (2) introduced diversified questioning approaches to increase variety, and (3) eliminated problems with conceptually similar solutions (e.g., those relying on the same underlying mathematical identity or solving trick) to ensure mathematical diversity.

This process yielded 86 high-quality seed puzzles, including 72 SMT-solvable and 14 pure formula/code-based (labeled as \texttt{A}-series).

% \begin{figure*}[tbp]
%   \centering
%   \includegraphics[width=0.7\linewidth]{figures/duplicate_report_hide_zero.png}
%   \caption{Distribution of duplicated instances across 32 seed puzzles. Only seed puzzles with at least one duplicate are shown.}
%   \label{fig:duplication}
% \end{figure*}

\subsection{Puzzle Generation and Deduplication} \label{sec:deduplication}
We utilized PuzzleClone to generate 1K variants for each seed puzzle, resulting in an initial set of 86K puzzle instances. 
For each generated variant, we assigned a unique identifier, its source puzzle, question type (\texttt{qtype}), and evaluation type (\texttt{eval\_type}). The \texttt{qtype} indicates the format of the question, such as multiple-choice, fill-in-the-blank, or short answer. The \texttt{eval\_type} specifies the answer structure and corresponding evaluation strategy, including formats like \texttt{numeral}, \texttt{nominal}, \texttt{option}, \texttt{ordered\_array}, and \texttt{unordered\_array}. Based on the assigned \texttt{qtype} and \texttt{eval\_type}, we designed custom prompt wrappers (Appendix \ref{sec:prompts}) and implemented appropriate evaluation operators.

% To ensure data quality, we applied a deduplication process to eliminate logically equivalent puzzles. Since each instance was generated independently and randomly, we compared their underlying \texttt{config} structures (Figure~\ref{fig:pipeline}D) rather than their surface-level text representations (Figure~\ref{fig:pipeline}E). 
% Since each instance was generated independently and randomly, we applied a deduplication process to eliminate logically equivalent puzzles by comparing their \texttt{configs} (Figure~\ref{fig:pipeline}D) rather than surface-level text (Figure~\ref{fig:pipeline}E).
% Specifically, we ignored nominal fields (e.g., \textit{names}) and treated value pools as unordered sets in puzzle \texttt{configs}, enabling us to identify near-duplicates even when textual details varied (such as wording or ordering differed).
Since instances were generated independently and randomly, we deduplicated puzzles by comparing their logical \texttt{configs} (Figure~\ref{fig:pipeline}D) rather than surface text (Figure~\ref{fig:pipeline}E).
Specifically, we ignored nominal fields (e.g., \textit{names}) and treated value pools as unordered sets, allowing detection of logically equivalent puzzles despite textual variations (e.g., differences in wording or ordering).

% After deduplication, we retained 83,657 unique puzzle instances. These puzzles cover diverse question types (e.g., multiple-choice and short-answer) and answer types (e.g., numeral values and arrays).
% Among the 86 seed puzzles, 32 had at least one duplicated variant.
% Notably, \texttt{28-exam} exhibited the highest duplication count, with 678 repeated instances. 
% Detailed statistics of the result dataset are provided in the Appendix \ref{sec:benchstats}.
% % As detailed in Appendix~\ref{app:duplication}, 
% Our analysis revealed that puzzles with limited randomizable space tend to result in higher duplication rates.
% % —a finding that aligns with our broader observations discussed in Section~\ref{sec:discussion}.

After deduplication, we retained 83,657 unique puzzle instances. Detailed statistics of the resulting dataset—including the distributions of question types, answer types, and duplication patterns—are provided in Appendix~\ref{sec:data_stats}. 
% Overall, puzzles with limited randomizable space tend to exhibit higher duplication rates.

\begin{figure}[b]
\centering
\includegraphics[width=\linewidth]{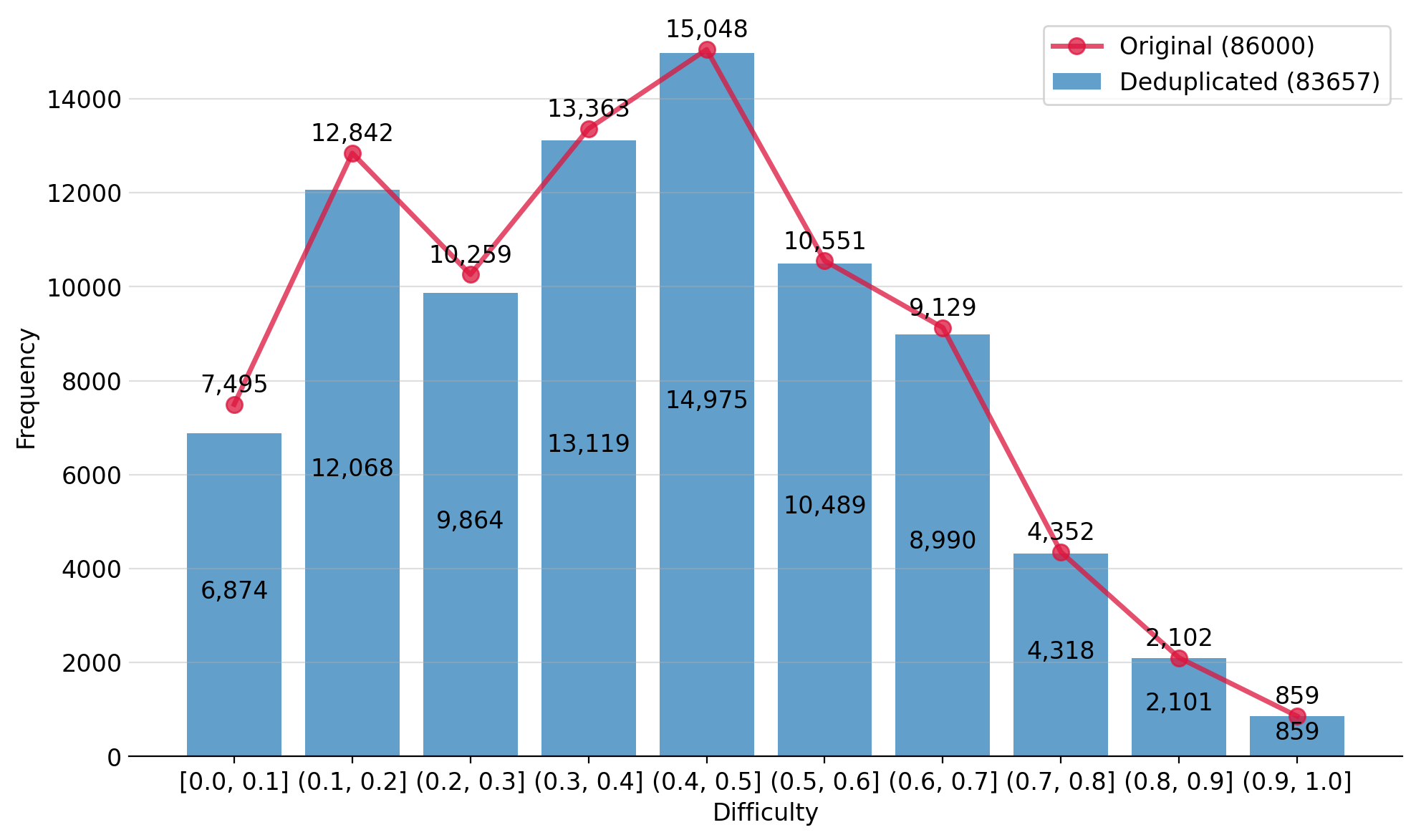}
\caption{Puzzle difficulty distribution before and after deduplication.}
\label{fig:difficulty_dist}
\end{figure}

\subsection{Difficulty Assessment and Analysis} \label{sec:diff_assess}

% To quantify puzzle complexity, we designed a heuristic difficulty score based on four features:
% \begin{compactitem}
% \item \texttt{sym\_num}: number of logical symbols.
% \item \texttt{cond\_num}: number of logical constraints.
% \item \texttt{desc\_len}: the character length of the problem description.
% \item \texttt{var\_scale}: a custom metric estimating the difficulty impact of variable domains. It is calculated as the average of normalized variable values, adjusted by their $diff\_factor$ (see Figure~\ref{fig:pipeline}B), which encodes the variable's correlation direction with problem difficulty.
% \end{compactitem}
% \item \texttt{var\_scale}: a custom metric estimating the difficulty impact of variable domains. It is derived by assigning each variable a difficulty factor (\texttt{diff\_factor}, see Figure~\ref{fig:pipeline}B), which encodes how increasing the variable’s value range affects problem difficulty: positively (+1), negatively (–1), or neutrally (0, by default).
% % indicating whether increasing the value range makes the problem harder (1), easier (-1), or has no effect (0).

% \[
% v_{adj} = 
% \begin{cases}
%     \hat{v} & \text{if } \text{diff\_factor} > 0 \\
%     1 - \hat{v} & \text{if } \text{diff\_factor} < 0 \\
%     \text{excluded} & \text{if } \text{diff\_factor} = 0
% \end{cases}
% \]

% \begin{equation}
% \label{eq:var_scale}
% \texttt{var\_scale}=
% \frac{\sum_{v:\,\texttt{d}\neq 0}
% \left(
% \texttt{d}\cdot
% \frac{v - v_{\min}}{v_{\max}-v_{\min}} 
% + \frac{1-\texttt{d}}{2}
% \right)}{\left|\{v: \texttt{d}\neq0\}\right|}
% \end{equation}

To quantify puzzle complexity, we designed a heuristic difficulty score based on four features: the number of logical symbols (\texttt{sym\_num}), the number of logical constraints (\texttt{cond\_num}), the length of problem description (\texttt{desc\_len}), and a variable-scale metric (\texttt{var\_scale}) capturing the difficulty impact of variable domains.
% , which is modulated by $diff\_factor$ (Figure~\ref{fig:pipeline}B).
% We provide details of the difficulty metric in Appendix \ref{sec:difficulty}. 
Details of the difficulty score computation and its empirical validation using model accuracy are given in Appendix~\ref{sec:difficulty}.

The difficulty distribution of the puzzles is shown in Figure~\ref{fig:difficulty_dist}. Hard puzzles are relatively rare due to increased likelihood of being unsolvable when the complexity and diversity (i.e., the variable values and constraint number) increase. Furthermore, duplicates were disproportionately found among easier puzzles due to their narrower randomization space. 

\subsection{Dataset Partitioning Strategy}

% To ensure balanced evaluation and effective model training for both supervised fine-tuning (SFT) and reinforcement learning (RL) tasks, 
To facilitate a comprehensive evaluation of popular LLMs while supporting both supervised fine-tuning (SFT) and reinforcement learning (RL) training, we implemented a stratified dataset partitioning strategy based on puzzle difficulty and seed origin.
% The splitting protocol was as follows: First, 
Specifically, puzzles were classified into \textit{normal} (difficulty $\leq$ 0.5) or \textit{hard} (difficulty $>$ 0.5) categories. For each seed puzzle, its variants in each category were randomly split into 90\% train and 10\% test. From the training set, 25 \textit{normal} and 25 \textit{hard} samples were selected per seed puzzle (falling back to normal in case of insufficient hard samples) to form the SFT set ($86\times25\times2 = 4300$). A smaller subset (5 \textit{normal} + 5 \textit{hard} per seed) was selected for RL validation (860 in total), with the remaining training samples assigned to RL training.

\begin{table}[h]
    \centering
    \scalebox{0.78}{
    \begin{tabular}{@{}l|c|ccc|l@{}}
        % \cmidrule(l){1-6}
        \toprule
         & \multicolumn{4}{c|}{Train} & \multirow{2}{*}{Test} \\
         & SFT & RL-Train & RL-Val & Total &  \\ \midrule
        \textit{Normal} & 2161  & 50738 & 430 & 51168 & 5730 \\
        \textit{Hard} & 2139  & 23616 & 430 & 24046 & 2713 \\ \midrule
        Sum & 4300 & 74354 & 860 & 75214 & 8443 \\ \bottomrule
    \end{tabular}}
    \caption{Dataset partitioning statistics.}
    \label{tbl:dataset_split}
    % \vspace{-0.9cm}  % 调整图片与上文的垂直距离
\end{table}

% \begin{table}[h]
%     \centering
%     \scalebox{0.8}{
%     \begin{tabular}{l|ccc|l}
%         % \cmidrule(l){1-6}
%         \toprule
%          & SFT & RL-Train & RL-Val & Test  \\ \midrule
%         \textit{Normal} & 2161  & 50738 & 430  & 5730 \\
%         \textit{Hard} & 2139  & 23616 & 430  & 2713 \\ \midrule
%         Sum & 4300 & 74354 & 860 & 8443 \\ \bottomrule
%     \end{tabular}}
%     \caption{Dataset partitioning statistics.}
%     % Each puzzle is first labeled as \textit{normal} or \textit{hard} based on a difficulty threshold of 0.5. For SFT, 25 examples from each difficulty band per seed puzzle are selected. For RL, 5 examples per band are selected for validation (RL-Val), and the rest are used for training (RL-Train).
%     \label{tbl:dataset_split}
%     % \vspace{-0.9cm}  % 调整图片与上文的垂直距离
% \end{table}

The final partitioning (Table~\ref{tbl:dataset_split}) ensures comprehensive coverage, with each set
% (SFT, RL-Train, RL-Val, and Test) 
containing puzzles derived from all 86 seed puzzles while maintaining difficulty stratification.
% drawing from all seed puzzles and preserving difficulty stratification.

\section{Experiments}
\subsection{Experimental Setup}
All model training and inference were conducted on NVIDIA H100 GPUs. We trained our models using an 8$\times$NVIDIA H100 GPU cluster, with Qwen2.5-7B-Instruct as the foundation model for all training experiments (detailed training paramter setting can be found in Appendix \ref{sec:training_param}). 
% During evaluation, open-source models were deployed on an 8$\times$NVIDIA H100 cluster. DeepSeek-R1-0528 was evaluated using a 16$\times$NVIDIA H100 setup. Proprietary models were evaluated via their official APIs. For fair comparison and reproducibility, the inference temperature was set to 0 throughout all evaluations.

\begin{table}[tbp]
    \centering
    \scalebox{0.7}{
    \begin{tabular}{l|ccc}
        \toprule
        \textbf{Model} & \textbf{Normal} & \textbf{Hard} & \textbf{Avg.} \\
        \midrule
        \multicolumn{4}{l}{\textbf{Proprietary Models}} \\
        ChatGPT-4o \cite{chatgpt-4o}                & 31.7 & 24.6 & 28.2 \\
        ChatGPT-o3 \cite{chatgpt-o3}                 & 87.1   & 83.4   & 85.3   \\
        ChatGPT-5 \cite{chatgpt-5}                                & \textbf{91.1}   & \textbf{86.3}   & \textbf{88.7}   \\
        Gemini-2.0-flash \cite{gemini20}           & 42.0   & 31.6   & 36.8   \\
        Gemini-2.5-pro \cite{gemini25}             & 75.8   & 67.2   & 71.5   \\
        Gemini-3-pro \cite{gemini3pro}           & 86.5   & 83.0   & 84.8   \\
        Claude-3.5-sonnet \cite{claude35}         & 37.6   & 27.4   & 32.5   \\
        Claude-4-sonnet \cite{claude4}           & 62.7   & 47.8   & 55.3   \\
        Seed1.6 \cite{seed-16}           & 87.8   & 82.4   & 85.1   \\
        \midrule
        \multicolumn{4}{l}{\textbf{GLM Series} \cite{GLM_Z1}}\\
        GLM-Z1-9B-0414             & 63.6 & 53.5 & 58.6 \\
        GLM-Z1-32B-0414            & 71.1 & 60.9 & 66.0 \\
        \midrule
        \multicolumn{4}{l}{\textbf{Qwen2.5 Series} \cite{Qwen25}} \\
        Qwen2.5-7B-Instruct        & 16.8 & 12.1 & 14.5 \\
        Qwen2.5-14B-Instruct       & 24.3 & 17.9 & 21.1 \\
        Qwen2.5-32B-Instruct       & 31.4 & 23.5 & 27.4 \\
        Qwen2.5-72B-Instruct       & 32.8 & 25.3 & 29.0 \\
        \midrule
        \multicolumn{4}{l}{\textbf{Qwen3 Series} \cite{Qwen3}}\\
        Qwen3-8B                   & 71.6 & 59.4 & 65.5 \\
        Qwen3-14B                  & 78.6 & 67.0 & 72.8 \\
        Qwen3-32B                  & 77.0 & 68.1 & 72.5 \\
        Qwen3-235B-A22B            & 82.9 & 73.8 & 78.3 \\
        \midrule
        \multicolumn{4}{l}{\textbf{DeepSeek Series} \cite{DeepSeek_R1_2025}} \\
        DeepSeek-R1-Distill-Qwen-14B   & 47.9 & 38.4 & 43.1 \\
        DeepSeek-R1-Distill-Qwen-32B   & 53.3 & 43.2 & 48.3 \\
        DeepSeek-R1-0528-Qwen3-8B      & 76.0 & 66.8 & 71.4 \\
        DeepSeek-R1-0528               & 88.7 & 82.6 & 85.6 \\
        \bottomrule
    \end{tabular}
    }
    \caption{Baseline performance on PC-83K}
    \label{tab:evaluation_result}
\end{table}

\begin{table*}[h]
    \centering
    % \scalebox{0.7}{
    \resizebox{\textwidth}{!}{ % 更精确的缩放控制
    \begin{tabular}{l|cc|c|cc|ccccc}
        \toprule
        \multirow{2}{*}{\textbf{Model}} & \multicolumn{2}{c|}{\textbf{PC-83K}$^{\dagger}$} & \multicolumn{1}{c|}{\textbf{PC-SL}} & \multicolumn{2}{c|}{\textbf{Logic Benchmarks}} & \multicolumn{5}{c}{\textbf{Mathematical Benchmarks}} \\
        \cmidrule(lr){2-3} \cmidrule(lr){5-6} \cmidrule(lr){7-11}
        & Normal & Hard & \textbf{35K}$^{\dagger}$ & SATBench & BBEH-mini & AIME24 & AIME25 & AMC2023 & MATH500 & OlympiadBench \\
        \midrule
        Qwen2.5-7B-Instruct & 16.8 & 12.1 & 9.6 & 51.6 & 11.3 & 13.3 & 6.7 & 52.5 & 75.2 & 41.0 \\
        SFT                 & 61.9   & 48.0   & 14.7  & \textbf{70.0} & 9.8 & \underline{20.0} & \underline{13.3} & \textbf{67.5} & \textbf{80.8} & 43.4 \\
        RL (PC-83K)                 & \textbf{71.0}   & \textbf{61.0}   & 15.2   & \underline{62.0} & \textbf{17.0} & 16.7 & \underline{13.3} & \underline{65.0} & 80.0 & \underline{44.4} \\
        \midrule
        SynLogic-7B$^{\ddagger}$ & -   & -   & -   & - & 8.0 & 10.0 & - & 55.0 & 71.8 & - \\
         RL (PC-SL-35K)                 & 22.0  & 14.3   & \textbf{55.3}   & 58.4 & \underline{16.5} & \textbf{23.3} & 10.0 & 62.5 & 79.8 & 42.4 \\
        RL (PC-83K+PC-SL-35K)                 & \underline{64.8}   & \underline{54.1}   & \underline{54.2}   & 57.2 & \textbf{17.0} & 16.7 & \textbf{16.7} & 60.0 & \underline{80.4} & \textbf{52.2} \\
        \bottomrule
    \end{tabular}
    }
    % \caption{Performance of models trained with PC-83K and/or PC-SL-35K train sets. Results on PC-83K and PC-SL-35K are reported on their respective held-out test sets.}
    % \caption{Performance of models trained on PC-83K and/or PC-SL-35K, evaluated on their respective held-out test sets and external logic and math benchmarks.}
    \caption{Performance of models trained on PC-83K and/or PC-SL-35K. $^{\dagger}$ indicates the corresponding held-out test sets. $^{\ddagger}$ denotes results reported in the SynLogic~\cite{liu2025synlogicsynthesizingverifiablereasoning} paper due to incompatible data formats.}
    \label{tab:smt_training_evaluation}
\end{table*}

\subsection{Baseline Results}
% 1. 2D: models, group by seed questions
% 2. 2D: models, group by config
We evaluate a wide range of models on the PC-83K test set, including both open-source and proprietary models. Similar to the training test, the test set is also divided into \textit{Normal} and \textit{Hard} subsets to reflect different levels of reasoning complexity. As shown in Table~\ref{tab:evaluation_result}, current LLMs remain limited in handling complex logical reasoning. 

Among the open-source models, the Qwen and GLM-Z1 series demonstrate strong performance trends as model size increases. Within the Qwen2.5 family, scores steadily improve from the 7B (14.5 average) to the 72B (29.0 average). Qwen3-8B achieves an average of 65.5, already outperforming many larger models from the previous generation. The Qwen3-14B and Qwen3-32B models achieve average scores of 72.8 and 72.5, respectively, while Qwen3-235B-A22B reaches 78.3.

The GLM-Z1 models also exhibit strong results. GLM-Z1-9B achieves 58.6 on average, while GLM-Z1-32B outperforms it with a score of 66.0. These results suggest the GLM-Z1 series is competitive with Qwen3 in the same parameter range.

The DeepSeek-R1-Distill models based on Qwen show moderate performance (43.1 and 48.3), while the full DeepSeek-R1-0528-Qwen3-8B model attains a strong 71.4. The most powerful open-source model DeepSeek-R1-0528 achieves an impressive 85.6 average score, surpassing all other open models and even some proprietary models.

Among the proprietary models, ChatGPT-5  leads (Avg. 88.7), ChatGPT-o3 (Avg. 85.3), and Seed1.6 (Avg. 85.1) are the top three, while Gemini 3 Pro (Avg. 84.8) shows the smallest \textit{Normal} to \textit{Hard} drop (-3.5), evidencing strong, stable reasoning. Gemini-2.5-pro ranks next (Avg. 71.5) with moderate robustness, while Claude-4-sonnet is mid-pack (Avg. 55.3) but degrades sharply on \textit{Hard}. Non-flagship models perform notably worse, with Gemini-2.0-flash at 36.8, ChatGPT-4o at 28.2, and Claude-3.5-sonnet at 32.5.

\subsection{Post Training}

To evaluate the quality of PC-83K, we perform post-training (SFT and RL) on Qwen2.5-7B-Instruct. During RL (GRPO \cite{GRPO}) training, we use Qwen2.5-7B-Instruct as the base model to directly compare the effectiveness of SFT and RL training stages. 
As shown in Table~\ref{tbl:dataset_split}, we use 74,354 training samples for the RL stage. For SFT, we first use DeepSeek-R1-0528 to generate thinking traces for 4,300 training samples to populate the reasoning context. We then apply rule-based filters to remove thinking traces with duplicated generations and multilingual reasoning content. After filtering, 3,574 samples remain for SFT.

To demonstrate its effectiveness in enhancing the logical reasoning capabilities of LLMs, we evaluate the model on standard logic benchmarks, including SATBench \cite{satbench} and BBEH-mini \cite{bbeh}, as well as mathematical benchmarks such as AIME24 \cite{aime2024}, AIME25 \cite{aime2025}, AMC2023 \cite{amc2023}, MATH500 \cite{math500}, and OlympiadBench \cite{olympiadbench}. These evaluations demonstrate the capability of our training data to enhance both logical reasoning and mathematical problem-solving.

The main results are presented in Table~\ref{tab:smt_training_evaluation}, demonstrating that model performance is significantly improved by training on the PC-83K dataset at both the SFT and RL stages.

RL achieves the most substantial improvement on the PC-83K test set, increasing the average accuracy over \textit{Normal} (16.8 to 71.0) and \textit{Hard} (12.1 to 61.0) puzzles from 14.5 (Qwen2.5-7B-Instruct) to 66.0, demonstrating strong in-distribution optimization.
On external logic and mathematical benchmarks, SFT slightly outperforms RL on average (43.5 vs.\ 42.6), while both significantly exceed the baseline (35.9).
Notably, the largest single gain is observed on SATBench, where SFT improves accuracy from 51.6 to 70.0 (+18.4).
However, SFT underperforms on BBEH-mini (9.8 vs.\ 11.3 for the baseline), whereas RL improves substantially to 17.0, suggesting that RL better captures deeper compositional logical reasoning.
Overall, these results highlight a trade-off between RL’s strong alignment with the target distribution and SFT’s slightly better cross-benchmark generalization, with particularly pronounced gains on logic-oriented tasks.

\subsection{Case Study: Generalizing SynLogic}

% In this section, we conducted a case study to show that PuzzleClone framework can be generalized to broader tasks with the dataset contributing to models' performances on unseen seeds. 
% We utilized our framework to reproduce and generalize SynLogic~\cite{liu2025synlogicsynthesizingverifiablereasoning}, a representative and diverse dataset of over 33K puzzles spanning 35 logical reasoning tasks. Specifically, we used these puzzles as seeds, curated DSLs to reproduce them, and successfully synthesized a new dataset of 35,829 puzzles (termed PC-SL-35K below) with PuzzleClone framework. The result dataset was divided into a training and test set on a 90-10 basis.

% We compared the quality of PC-SL-35K and and the original SynLogic dataset with two studies. First, we evaluated the accuracy of diverse models on the PC-SL-35K test set. As shown in Table ~\ref{tab:smt_training_evaluation}, the results show that the difficulty of puzzles in both datasets are comparable. In addition, when tested on PC-SL-35K, the performance of models trained on PC-83K by RL and SFT rose from 9.6 to 15.2 and 14.7 respectively, showing that the models' capabilities can be generalized to puzzles synthesized from unseen seeds. 
% Meanwhile, we used the training set of PC-SL-35K for SFT and RL on the Qwen2.5-7B-Instruct model. \todo{Results to be updated} 

To demonstrate the practicality and generalizability of PuzzleClone, we conduct a case study by reproducing and extending SynLogic~\cite{liu2025synlogicsynthesizingverifiablereasoning}, a representative dataset of over 33K puzzles spanning 35 logical reasoning tasks.
Specifically, we treat SynLogic puzzles as seeds and curate DSL specs to synthesize a new dataset of 35,829 puzzles with PuzzleClone, termed PC-SL-35K, which is split into training and test sets using a 90/10 ratio.
% The resulting dataset is split into training and test sets on a 90-10 basis.
% We perform two RL experiments: one trained solely on PC-SL-35K, and another trained on the combination of PC-83K and PC-SL-35K. The performance of models summarized in Table~\ref{tab:smt_training_evaluation}.
We perform two RL experiments, trained on PC-SL-35K and on PC-83K+PC-SL-35K, respectively, with results summarized in Table~\ref{tab:smt_training_evaluation}.

Our analysis yields four key observations.
% First, Qwen2.5-7B-Instruct achieves an accuracy of 9.0 on the original SynLogic benchmark and 9.6 on the PC-SL-35K test set, indicating that the reproduced dataset exhibits a difficulty level comparable to the original SynLogic data.
First, Qwen2.5-7B-Instruct achieves comparable accuracy on the original SynLogic benchmark (9.0) and the PC-SL-35K test set (9.6), indicating similar difficulty levels across the two datasets.
Second, RL (PC-SL-35K) consistently outperforms the SynLogic-7B baseline on external benchmarks, demonstrating that PuzzleClone-synthesized data is at least as effective as SynLogic data in enhancing model reasoning performance.
Third, while RL (PC-83K+PC-SL-35K) generally outperforms RL (PC-SL-35K), it slightly underperforms RL (PC-83K) on some benchmarks, which we attribute to differences in training progress (600 vs.\ 900 steps).
% We attribute this discrepancy to differences in training progress: the combined model is evaluated at an earlier checkpoint (600 steps) due to time constraints, whereas the PC-83K model is evaluated at 900 steps.
% We plan to continue training the combined model and reassess its performance in future work.
Fourth, both SFT and RL (PC-83K) models achieve higher accuracy on the PC-SL-35K test set than Qwen2.5-7B-Instruct, indicating improved generalization to unseen seed puzzles and further validating the effectiveness of PuzzleClone.
% showing that training on PuzzleClone-generated data improves generalization to unseen seed puzzles, further validating the effectiveness of the PuzzleClone framework.

\section{Related Work}

\subsection{Data generation}

% The growing requirement of robust and powerful  large language models have led to a gigantic need for high-quality and diverse data. Due to the limited amount of publicly available datasets and manually curated ones, various data generation approaches have been designed, among which \textit{data annotation}, \textit{data augmentation}, and \textit{data synthesis} are prevalent methods. 

Research on data generation mainly focus on \textit{data synthesis} and \textit{data augmentation}~\cite{Wang2024}.

Data synthesis approaches broadly start by extracting instructions from existing data sources or synthesizing them with rule-based or model-based synthesizers~\cite{maiya2025explaining, ding2023enhancing, xu2024magpie}. Next, responses will be generated for each instruction instance, either manually or automatically. 
Machine learning has been widely adopted to accelerate the annotation process of various types of data~\cite{zhang2021survey, lu2023machine}, despite their limitations in efficiency, data consistency, and domain knowledge~\cite{tan-etal-2024-large}. These limitations are addressed by modern LLMs, whose reasoning abilities make them powerful universal annotators~\cite{zhang2021survey, lu2023machine, csanady2024llambert, zhang2023llmaaa}. 
Tan et al. provide a detailed survey of these methods~\cite{tan-etal-2024-large}.
However, expert knowledge remains essential for accurate labels. 
Consequently, semi-automatic approaches have emerged, utilizing human-AI collaborative tools and co-labeling workflows to enhance efficiency and accuracy~\cite{li2023coannotating, wang2024model, coral}.

By contrast, data augmentation focuses on adapting existing data items, called seeds, to new instances with similar structures. In natural language processing, traditional approaches mainly focus on paraphrasing the instructions, leveraging approaches like rule-based transformation and language models~\cite{feng-etal-2021-survey, xu2024wizardlm, sugiyama2019data, bayer2022survey}. With the advent of LLMs, modern approaches have explored involving LLMs throughout the pipeline~\cite{Wang2024}, allowing LLMs to automatically select promising seeds, generate data and self-improve their reasoning, fine-tune the instructions, and perform evaluation-based filtering to enhance data quality~\cite{Lu2024, Shah2024, huang2024datagen}. Some works further involve human validators or LLM-generated scripts in symbolic languages for better accuracy~\cite{Leang2025, Shah2024}. However, there remains a need for robust and scalable strategies that enable end-to-end verification and fine-grained controllability throughout the pipeline.
\subsection{Benchmarks for logical reasoning}
Numerous public datasets for mathematical and logical reasoning have been collected through academic challenges and crowdsourcing~\cite{hendrycksmath2021, math-ai_aime25}. Curated datasets, however, mostly focus on specific classes of puzzles. For instance, numerous studies generate equation systems or deductive, inductive, and abductive puzzles by combining unit equations or basic logical clauses  while crafting natural language instructions using templates or language models~\cite{chen2025justlogic, parmar2024logicbench, luo2023towards, wei2025satbench, mirzadeh2024gsm}. However, it becomes increasingly difficult to map the abstract logical model to a real-world scenario when it becomes difficult, which in turn jeopardizes the articulation of the instructions. By contrast, another line of research start from classical complicated and challenging mathematical puzzles such as sudoku, hitori, and crosswords, and curate a number of similar puzzles simply through rule-based methods~\cite{gui2024logicgame, mittal_puzzlebench_2024, chen_lr2bench_2025, chen_finereason_2025, li2025logic, liu2025combibench, liu2025synlogicsynthesizingverifiablereasoning}. However, a universal framework capable of generating various challenging puzzles is still absent, leading us to design the PuzzleClone framework. 

\section{Discussion}
% Implications
In the process of designing and implementing the PuzzleClone framework, as well as synthesizing variants from seed puzzles, we encountered several key insights and reflections worth discussing.

% \subsection{Diversity in Puzzle and Evaluation Types}
\textbf{Diverse Question and Evaluation Types:}
Unlike some existing datasets that typically focus on a single task format, our dataset features a wide range of puzzle types and answer formats (Appendix~\ref{sec:data_stats}).
Notably, a single puzzle in our dataset may contain multiple sub-questions, each with its own \texttt{qtype} and \texttt{eval\_type}, which further enhances the overall richness and versatility of the dataset.
Such diversity enables more comprehensive benchmarking, supports evaluation of specific reasoning capabilities, and promotes model robustness across heterogeneous formats—better reflecting real-world problem-solving scenarios.
% This diversity brings several benefits. First, it enables comprehensive benchmarking across a broader range of reasoning and answer-evaluation challenges, providing a more realistic testbed for general-purpose solvers. Second, it facilitates targeted evaluation: researchers can isolate performance on specific reasoning modes or output types. Finally, it promotes model robustness, as systems trained or tested on this dataset must handle heterogeneous formats, reflecting the variety of real-world tasks encountered in educational, logical, or cognitive settings.

% \subsection{Forward vs. Backward Specification Strategies}
\textbf{Specification Strategies:}
We observed two distinct approaches to puzzle synthesis based on the specification: forward and backward generation. In the forward strategy (e.g., specs for Figure~\ref{fig:pipeline}, \texttt{15-tabletennis}, and \texttt{23-product}), we first define random domains for variables and use symbolic constraints to allow the solver to search for feasible solutions. In contrast, the backward strategy (e.g., specs for \texttt{7-age} and \texttt{11-wine}) involves randomly generating a feasible solution first, then building symbolic constraints around it to verify its correctness and uniqueness using the solver.
The backward approach significantly improves generation efficiency by avoiding unsolvable instances due to random variable combinations. Therefore, we recommend using the backward strategy for most specification designs.

% \subsection{Specification Design Requires Semantic Awareness}
\textbf{Hyperparameter Sensitivity:}
While writing specifications, careful tuning of hyperparameters such as variable domains, constraint templates, and their count ranges is crucial to maintain the semantic soundness of generated puzzles. For example, in \texttt{2-graduation}, where $m$ out of $n$ students are selected under constraints, it is necessary to enforce $0 < m < n$. However, making $m$ too small or too large ($m\rightarrow0$ or $m\rightarrow n$) reduces puzzle difficulty. Thus, domain ranges should be both logically valid and aligned with the intended cognitive challenge.

% \subsection{On the (In)accuracy of Difficulty Estimation}

% Yanwei[20251214]: 我注释掉了这个limitation，因为Z3不再重要了。
% \textbf{Expressiveness for Optimization Tasks:}
% Although PuzzleClone supports synthesis of puzzles involving optimization objectives, it is limited to linear optimization due to the capabilities of Z3. This restricts the types of puzzles our system can synthesize. 
% Extending support to nonlinear optimization remains an open challenge.

% \subsection{Distribution Bias in Generated Puzzle Difficulty}

\begin{figure}[tbp]
  \centering
  \includegraphics[width=\linewidth]{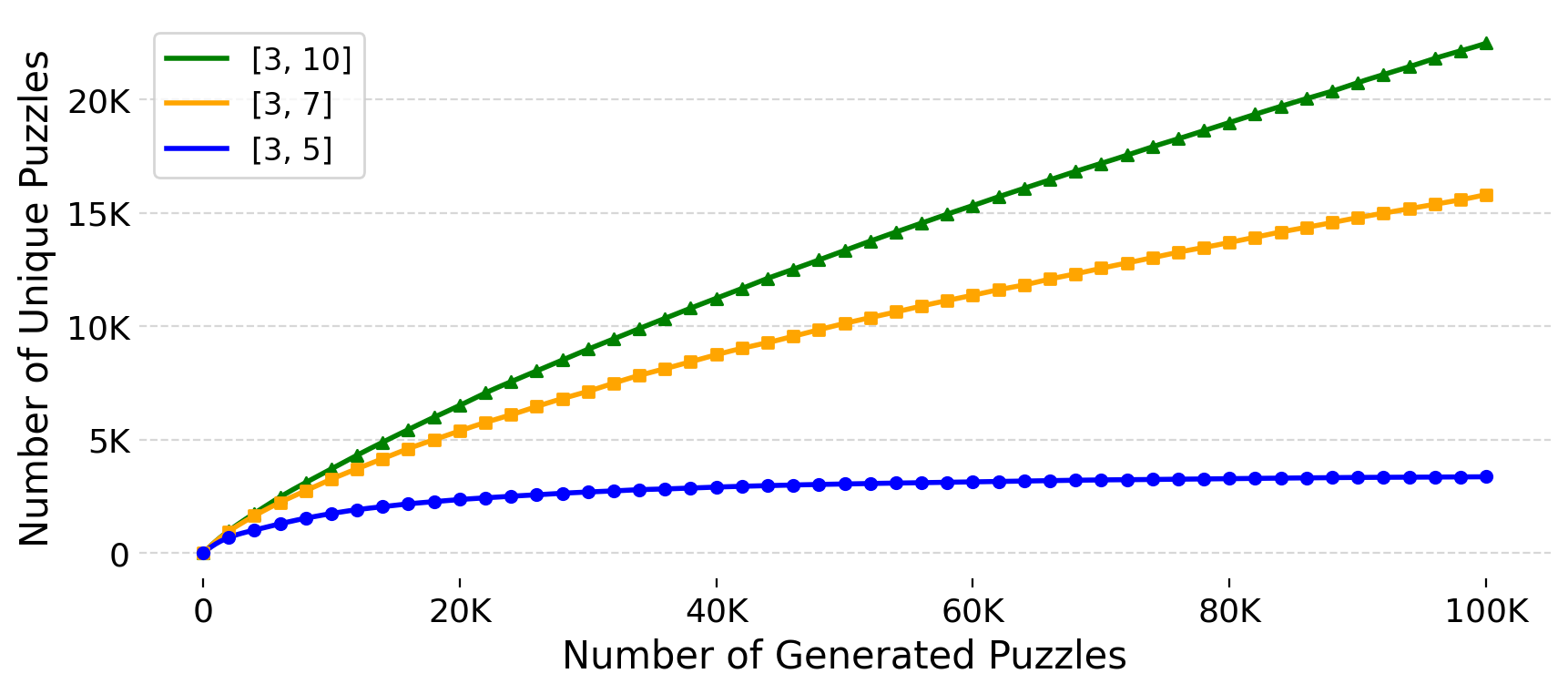}
  \caption{Number of valid puzzles (after de-duplication) synthesized from the seed puzzle \texttt{9-vase} using three different \texttt{p\_num} domain settings.}
  % : \texttt{[3,5]}, \texttt{[3,7]}, and \texttt{[3,10]}. The curves show how broader domain settings support more diverse puzzle synthesis, with smaller domains reaching saturation earlier.
  \label{fig:valid-comparison}
  % \vspace{-0.4cm}  % 调整图片与上文的垂直距离
\end{figure}

% \subsection{Finite Limit of Synthesizable Valid Puzzles}
\textbf{Upper Bound of Synthesizable Puzzles:}
While PuzzleClone can synthesize puzzles at scale, each specification inherently has a finite upper bound on the number of valid (i.e., non-duplicate) puzzles it can generate. This bound is determined by the size of the puzzle space defined in the specification.
As shown in Figure~\ref{fig:valid-comparison}, for \texttt{9-vase}, we synthesized variants using three specs with decreasing \texttt{p\_num} domains: {[3,10]}, {[3,7]}, and {[3,5]}. 
% Despite generating 100K puzzles in each case, de-duplication revealed early saturation in the first two settings (at roughly 4K and 8K respectively), whereas the third continued growing. 
After generating 100K puzzles for each setting and performing de-duplication, we observed that {[3,5]} quickly saturated at around 3.3K unique puzzles, while {[3,10]} continued to grow steadily. The {[3,7]} setting fell in between and showed signs of approaching saturation.
This demonstrates that a larger solution space allows the synthesis of more diverse valid puzzles.

% \subsection{Avoiding Redundant Information Between Problem and Conditions}
\textbf{Preventing Redundancy in Puzzle Statements:}
It is critical to ensure that the question and its constraints do not inadvertently leak the answer. For instance, in \texttt{23-product}, if a condition states that a specific product is in position $i$ and the question also asks about the position of that product, the answer becomes trivially inferable without reasoning. Such overlap undermines the puzzle’s intent. Specifications should explicitly avoid this redundancy to preserve reasoning difficulty.

\textbf{Advanced Features for Practical Synthesis:}
To enhance expressiveness and practical utility, PuzzleClone supports several advanced features:

\begin{compactitem}
% \begin{itemize}
% \item \textbf{Custom Conditions:} Allows inter-variable/constraint dependencies (e.g., $A + B = 10$) beyond independent random sampling.
\item \textit{Coupled Logic:} Supports joint control of variables and constraints (e.g., $A + B = 10$), enabling fine-grained coordination beyond independent random sampling.
\item \textit{Custom Operators:} Users can define new functions or import scripts, compensating for limitations in built-in operators.
\item \textit{Dynamic Rephrasing:} PuzzleClone supports modifying the textual surface of puzzles (see Appendix),
% (see tutorial\footnote{https://puzzleclone.github.io/PuzzleClone/tutorial/configurations}),
enabling translation into multiple languages or thematic adaptation, thereby increasing data diversity.
% \end{itemize}
\end{compactitem}

\section{Conclusion}

We present PuzzleClone, a DSL-driven data synthesis framework and a diverse and challenging dataset containing over 83K puzzles.
% Future work includes leveraging dynamic rephrasing for multilingual adaptation, exploring LLM-driven constraint enrichment to increase puzzle complexity, and automating specification construction to streamline puzzle generation.
Experiments show that existing state-of-the-art models
% including DeepSeek and Qwen series, 
face significant challenges on this benchmark. Furthermore, SFT and RL applied to Qwen2.5-7B-Instruct show substantial gains in mathematical reasoning, validating the effectiveness of our approach.

\section*{Limitations}

% While our PuzzleClone framework demonstrates promising results in generating diverse and challenging mathematical puzzles, several limitations remain. 
% We highlight the main challenges below and outline potential directions for future work. 

\textbf{Challenges in measuring puzzle difficulty accurately:} 
Although the experiment in Appendix~\ref{sec:difficulty} demonstrates the validity of the difficult metric, it has two main limitations. 
First, the \texttt{var\_scale} metric relies on a simple \texttt{diff\_factor} with only three discrete values (+1, 0, –1) to indicate correlation with puzzle difficulty, which cannot capture more nuanced relationships. 
Second, not all difficulty-variable relationships are monotonic; for example, increasing a variable’s value may initially make a puzzle harder but later easier.
Future work could explore more expressive difficulty metrics that better capture non-linear and variable-specific effects.

\textbf{Biased puzzle difficulty distribution:} 
When variable domains are large and constraints are complex, the probability of randomly generating solvable instances diminishes. This leads to a distributional bias in synthesized puzzles (see Figure~\ref{fig:difficulty_dist}): easier puzzles (associated with simpler parameter combinations) are overrepresented, while harder ones are under-sampled. 
Future research could design targeted generation procedures to ensure adequate representation of challenging puzzles.

\textbf{Human effort in specification authoring:} 
Authoring DSL specifications from seed puzzles currently requires substantial manual effort, including decomposing and rephrasing puzzles, brainstorming new variants and constraints to enhance puzzle diversity and complexity, and encoding them into DSL specifications. 
This labor-intensive process limits scalability.
A promising future direction is to leverage LLMs to automate these steps with minimal human intervention.

\bibliography{custom}

@article{Shah2024,
archivePrefix = {arXiv},
arxivId = {2407.21009},
author = {Shah, Vedant and Yu, Dingli and Lyu, Kaifeng and Park, Simon and Ke, Nan Rosemary and Mozer, Michael and Bengio, Yoshua and Arora, Sanjeev and Goyal, Anirudh},
eprint = {2407.21009},
title = {{AI-Assisted Generation of Difficult Math Questions}},
url = {http://arxiv.org/abs/2407.21009},
year = {2024}
}

@book{Wang2024,
archivePrefix = {arXiv},
arxivId = {2410.12896},
author = {Wang, Ke and Zhu, Jiahui and Ren, Minjie and Liu, Zeming and Li, Shiwei and Zhang, Zongye and Zhang, Chenkai and Wu, Xiaoyu and Zhan, Qiqi and Liu, Qingjie and Wang, Yunhong},
eprint = {2410.12896},
number = {1},
publisher = {Association for Computing Machinery},
title = {{A Survey on Data Synthesis and Augmentation for Large Language Models}},
url = {http://arxiv.org/abs/2410.12896},
volume = {1},
year = {2024}
}

@inproceedings{Lu2024,
address = {Stroudsburg, PA, USA},
archivePrefix = {arXiv},
arxivId = {2412.11936},
author = {Lu, Zimu and Zhou, Aojun and Ren, Houxing and Wang, Ke and Shi, Weikang and Pan, Junting and Zhan, Mingjie and Li, Hongsheng},
booktitle = {Proceedings of the 62nd Annual Meeting of the Association for Computational Linguistics (Volume 1: Long Papers)},
doi = {10.18653/v1/2024.acl-long.151},
eprint = {2412.11936},
pages = {2732--2747},
publisher = {Association for Computational Linguistics},
title = {{MathGenie: Generating Synthetic Data with Question Back-translation for Enhancing Mathematical Reasoning of LLMs}},
url = {http://arxiv.org/abs/2412.11936 https://aclanthology.org/2024.acl-long.151},
year = {2024}
}

@article{Leang2025,
archivePrefix = {arXiv},
arxivId = {2502.13137},
author = {Leang, Joshua Ong Jun and Hong, Giwon and Li, Wenda and Cohen, Shay B.},
eprint = {2502.13137},
title = {{Theorem Prover as a Judge for Synthetic Data Generation}},
url = {http://arxiv.org/abs/2502.13137},
year = {2025}
}

@inproceedings{feng-etal-2021-survey,
title = "A Survey of Data Augmentation Approaches for {NLP}",
author = "Feng, Steven Y.  and
  Gangal, Varun  and
  Wei, Jason  and
  Chandar, Sarath  and
  Vosoughi, Soroush  and
  Mitamura, Teruko  and
  Hovy, Eduard",
editor = "Zong, Chengqing  and
  Xia, Fei  and
  Li, Wenjie  and
  Navigli, Roberto",
booktitle = "Findings of the Association for Computational Linguistics: ACL-IJCNLP 2021",
month = aug,
year = "2021",
address = "Online",
publisher = "Association for Computational Linguistics",
url = "https://aclanthology.org/2021.findings-acl.84/",
doi = "10.18653/v1/2021.findings-acl.84",
pages = "968--988"
}

@inproceedings{tan-etal-2024-large,
    title = "Large Language Models for Data Annotation and Synthesis: A Survey",
    author = "Tan, Zhen  and
      Li, Dawei  and
      Wang, Song  and
      Beigi, Alimohammad  and
      Jiang, Bohan  and
      Bhattacharjee, Amrita  and
      Karami, Mansooreh  and
      Li, Jundong  and
      Cheng, Lu  and
      Liu, Huan",
    editor = "Al-Onaizan, Yaser  and
      Bansal, Mohit  and
      Chen, Yun-Nung",
    booktitle = "Proceedings of the 2024 Conference on Empirical Methods in Natural Language Processing",
    month = nov,
    year = "2024",
    address = "Miami, Florida, USA",
    publisher = "Association for Computational Linguistics",
    url = "https://aclanthology.org/2024.emnlp-main.54/",
    doi = "10.18653/v1/2024.emnlp-main.54",
    pages = "930--957"
}

@inproceedings{z3,
author = {De Moura, Leonardo and Bj\o{}rner, Nikolaj},
title = {Z3: an efficient SMT solver},
year = {2008},
isbn = {3540787992},
publisher = {Springer-Verlag},
address = {Berlin, Heidelberg},
booktitle = {Proceedings of the Theory and Practice of Software, 14th International Conference on Tools and Algorithms for the Construction and Analysis of Systems},
pages = {337–340},
numpages = {4},
location = {Budapest, Hungary},
series = {TACAS'08/ETAPS'08}
}

@inproceedings{pan-etal-2023-logic,
    title = "Logic-{LM}: Empowering Large Language Models with Symbolic Solvers for Faithful Logical Reasoning",
    author = "Pan, Liangming  and
      Albalak, Alon  and
      Wang, Xinyi  and
      Wang, William",
    editor = "Bouamor, Houda  and
      Pino, Juan  and
      Bali, Kalika",
    booktitle = "Findings of the Association for Computational Linguistics: EMNLP 2023",
    month = dec,
    year = "2023",
    address = "Singapore",
    publisher = "Association for Computational Linguistics",
    url = "https://aclanthology.org/2023.findings-emnlp.248/",
    doi = "10.18653/v1/2023.findings-emnlp.248",
    pages = "3806--3824"
}

@book{clocksin2003programming,
  title={Programming in PROLOG},
  author={Clocksin, William F and Mellish, Christopher S},
  year={2003},
  publisher={Springer Science \& Business Media}
}

@inproceedings{barrett2010smt,
  title={The smt-lib standard: Version 2.0},
  author={Barrett, Clark and Stump, Aaron and Tinelli, Cesare and others},
  booktitle={Proceedings of the 8th international workshop on satisfiability modulo theories (Edinburgh, UK)},
  volume={13},
  pages={14},
  year={2010}
}

@article{li2025logic,
  title={Logic-of-Thought: Empowering Large Language Models with Logic Programs for Solving Puzzles in Natural Language},
  author={Li, Naiqi and Liu, Peiyuan and Liu, Zheng and Dai, Tao and Jiang, Yong and Xia, Shu-Tao},
  journal={arXiv preprint arXiv:2505.16114},
  year={2025}
}

@article{liu2025combibench,
  title={CombiBench: Benchmarking LLM capability for combinatorial mathematics},
  author={Liu, Junqi and Lin, Xiaohan and Bayer, Jonas and Dillies, Yael and Jiang, Weijie and Liang, Xiaodan and Soletskyi, Roman and Wang, Haiming and Xie, Yunzhou and Xiong, Beibei and others},
  journal={arXiv preprint arXiv:2505.03171},
  year={2025}
}

@article{zhang2021survey,
  title={A survey on machine learning techniques for auto labeling of video, audio, and text data},
  author={Zhang, Shikun and Jafari, Omid and Nagarkar, Parth},
  journal={arXiv preprint arXiv:2109.03784},
  year={2021}
}

@article{csanady2024llambert,
  title={LlamBERT: Large-scale low-cost data annotation in NLP},
  author={Csan{\'a}dy, B{\'a}lint and Muzsai, Lajos and Vedres, P{\'e}ter and N{\'a}dasdy, Zolt{\'a}n and Luk{\'a}cs, Andr{\'a}s},
  journal={arXiv preprint arXiv:2403.15938},
  year={2024}
}

@article{zhang2023llmaaa,
  title={Llmaaa: Making large language models as active annotators},
  author={Zhang, Ruoyu and Li, Yanzeng and Ma, Yongliang and Zhou, Ming and Zou, Lei},
  journal={arXiv preprint arXiv:2310.19596},
  year={2023}
}

@article{li2023coannotating,
  title={Coannotating: Uncertainty-guided work allocation between human and large language models for data annotation},
  author={Li, Minzhi and Shi, Taiwei and Ziems, Caleb and Kan, Min-Yen and Chen, Nancy F and Liu, Zhengyuan and Yang, Diyi},
  journal={arXiv preprint arXiv:2310.15638},
  year={2023}
}

@article{wang2024model,
  title={Model-in-the-Loop (MILO): Accelerating Multimodal AI Data Annotation with LLMs},
  author={Wang, Yifan and Stevens, David and Shah, Pranay and Jiang, Wenwen and Liu, Miao and Chen, Xu and Kuo, Robert and Li, Na and Gong, Boying and Lee, Daniel and others},
  journal={arXiv preprint arXiv:2409.10702},
  year={2024}
}

@article{coral,
author = {Zhu, Zhen and Wang, Yibo and Yang, Shouqing and Long, Lin and Wu, Runze and Tang, Xiu and Zhao, Junbo and Wang, Haobo},
title = {CORAL: Collaborative Automatic Labeling System Based on Large Language Models},
year = {2024},
issue_date = {August 2024},
publisher = {VLDB Endowment},
volume = {17},
number = {12},
issn = {2150-8097},
url = {https://doi.org/10.14778/3685800.3685885},
doi = {10.14778/3685800.3685885},
journal = {Proc. VLDB Endow.},
month = aug,
pages = {4401–4404},
numpages = {4}
}

@article{lu2023machine,
  title={Machine learning for synthetic data generation: a review},
  author={Lu, Yingzhou and Shen, Minjie and Wang, Huazheng and Wang, Xiao and van Rechem, Capucine and Fu, Tianfan and Wei, Wenqi},
  journal={arXiv preprint arXiv:2302.04062},
  year={2023}
}

@inproceedings{sugiyama2019data,
  title={Data augmentation using back-translation for context-aware neural machine translation},
  author={Sugiyama, Amane and Yoshinaga, Naoki},
  booktitle={Proceedings of the fourth workshop on discourse in machine translation (DiscoMT 2019)},
  pages={35--44},
  year={2019}
}

@inproceedings{xu2024wizardlm,
  title={WizardLM: Empowering large pre-trained language models to follow complex instructions},
  author={Xu, Can and Sun, Qingfeng and Zheng, Kai and Geng, Xiubo and Zhao, Pu and Feng, Jiazhan and Tao, Chongyang and Lin, Qingwei and Jiang, Daxin},
  booktitle={The Twelfth International Conference on Learning Representations},
  year={2024}
}

@article{bayer2022survey,
  title={A survey on data augmentation for text classification},
  author={Bayer, Markus and Kaufhold, Marc-Andr{\'e} and Reuter, Christian},
  journal={ACM Computing Surveys},
  volume={55},
  number={7},
  pages={1--39},
  year={2022},
  publisher={ACM New York, NY}
}

@article{maiya2025explaining,
  title={Explaining Puzzle Solutions in Natural Language: An Exploratory Study on 6x6 Sudoku},
  author={Maiya, Anirudh and Alghamdi, Razan and Pacheco, Maria Leonor and Trivedi, Ashutosh and Somenzi, Fabio},
  journal={arXiv preprint arXiv:2505.15993},
  year={2025}
}

@article{ding2023enhancing,
  title={Enhancing chat language models by scaling high-quality instructional conversations},
  author={Ding, Ning and Chen, Yulin and Xu, Bokai and Qin, Yujia and Zheng, Zhi and Hu, Shengding and Liu, Zhiyuan and Sun, Maosong and Zhou, Bowen},
  journal={arXiv preprint arXiv:2305.14233},
  year={2023}
}

@article{xu2024magpie,
  title={Magpie: Alignment data synthesis from scratch by prompting aligned llms with nothing},
  author={Xu, Zhangchen and Jiang, Fengqing and Niu, Luyao and Deng, Yuntian and Poovendran, Radha and Choi, Yejin and Lin, Bill Yuchen},
  journal={arXiv preprint arXiv:2406.08464},
  year={2024}
}

@article{gui2024logicgame,
  title={Logicgame: Benchmarking rule-based reasoning abilities of large language models},
  author={Gui, Jiayi and Liu, Yiming and Cheng, Jiale and Gu, Xiaotao and Liu, Xiao and Wang, Hongning and Dong, Yuxiao and Tang, Jie and Huang, Minlie},
  journal={arXiv preprint arXiv:2408.15778},
  year={2024}
}

@article{chen2025justlogic,
  title={JustLogic: A Comprehensive Benchmark for Evaluating Deductive Reasoning in Large Language Models},
  author={Chen, Michael K and Zhang, Xikun and Tao, Dacheng},
  journal={arXiv preprint arXiv:2501.14851},
  year={2025}
}

@article{parmar2024logicbench,
  title={LogicBench: Towards systematic evaluation of logical reasoning ability of large language models},
  author={Parmar, Mihir and Patel, Nisarg and Varshney, Neeraj and Nakamura, Mutsumi and Luo, Man and Mashetty, Santosh and Mitra, Arindam and Baral, Chitta},
  journal={arXiv preprint arXiv:2404.15522},
  year={2024}
}

@article{luo2023towards,
  title={Towards logiglue: A brief survey and a benchmark for analyzing logical reasoning capabilities of language models},
  author={Luo, Man and Kumbhar, Shrinidhi and Parmar, Mihir and Varshney, Neeraj and Banerjee, Pratyay and Aditya, Somak and Baral, Chitta and others},
  journal={arXiv preprint arXiv:2310.00836},
  year={2023}
}

@article{wei2025satbench,
  title={SATBench: Benchmarking LLMs' Logical Reasoning via Automated Puzzle Generation from SAT Formulas},
  author={Wei, Anjiang and Wu, Yuheng and Wan, Yingjia and Suresh, Tarun and Tan, Huanmi and Zhou, Zhanke and Koyejo, Sanmi and Wang, Ke and Aiken, Alex},
  journal={arXiv preprint arXiv:2505.14615},
  year={2025}
}

@article{hendrycksmath2021,
  title={Measuring Mathematical Problem Solving With the MATH Dataset},
  author={Dan Hendrycks and Collin Burns and Saurav Kadavath and Akul Arora and Steven Basart and Eric Tang and Dawn Song and Jacob Steinhardt},
  journal={NeurIPS},
  year={2021}
}

@misc{math-ai_aime25,
  author       = {{Math-AI}},
  title        = {AIME25 Dataset},
  year         = {2025},
  url          = {https://huggingface.co/datasets/math-ai/aime25},
  note         = {Accessed: 2025-07-01}
}

@misc{chen_lr2bench_2025,
	title = {{LR\textsuperscript{2}Bench}: {Evaluating} {Long}-chain {Reflective} {Reasoning} {Capabilities} of {Large} {Language} {Models} via {Constraint} {Satisfaction} {Problems}},
	shorttitle = {{LR}\${\textasciicircum}2\${Bench}},
	url = {http://arxiv.org/abs/2502.17848},
	doi = {10.48550/arXiv.2502.17848},
	language = {en},
	urldate = {2025-06-13},
	publisher = {arXiv},
	author = {Chen, Jianghao and Wei, Zhenlin and Ren, Zhenjiang and Li, Ziyong and Zhang, Jiajun},
	month = apr,
	year = {2025},
	note = {arXiv:2502.17848 [cs]},
}

@misc{mittal_puzzlebench_2024,
	title = {{PuzzleBench}: {Can} {LLMs} {Solve} {Challenging} {First}-{Order} {Combinatorial} {Reasoning} {Problems}?},
	shorttitle = {{PuzzleBench}},
	url = {http://arxiv.org/abs/2402.02611},
	doi = {10.48550/arXiv.2402.02611},
	urldate = {2025-06-14},
	publisher = {arXiv},
	author = {Mittal, Chinmay and Kartik, Krishna and Mausam and Singla, Parag},
	month = feb,
	year = {2024},
	note = {arXiv:2402.02611 [cs]
version: 2},
}

@misc{chen_finereason_2025,
	title = {{FINEREASON}: {Evaluating} and {Improving} {LLMs}' {Deliberate} {Reasoning} through {Reflective} {Puzzle} {Solving}},
	shorttitle = {{FINEREASON}},
	url = {http://arxiv.org/abs/2502.20238},
	doi = {10.48550/arXiv.2502.20238},
	urldate = {2025-06-16},
	publisher = {arXiv},
	author = {Chen, Guizhen and Xu, Weiwen and Zhang, Hao and Chan, Hou Pong and Liu, Chaoqun and Bing, Lidong and Zhao, Deli and Luu, Anh Tuan and Rong, Yu},
	month = jun,
	year = {2025},
	note = {arXiv:2502.20238 [cs]},
}

@article{mirzadeh2024gsm,
  title={Gsm-symbolic: Understanding the limitations of mathematical reasoning in large language models},
  author={Mirzadeh, Iman and Alizadeh, Keivan and Shahrokhi, Hooman and Tuzel, Oncel and Bengio, Samy and Farajtabar, Mehrdad},
  journal={arXiv preprint arXiv:2410.05229},
  year={2024}
}

@inproceedings{huang2024datagen,
  title={Datagen: Unified synthetic dataset generation via large language models},
  author={Huang, Yue and Wu, Siyuan and Gao, Chujie and Chen, Dongping and Zhang, Qihui and Wan, Yao and Zhou, Tianyi and Xiao, Chaowei and Gao, Jianfeng and Sun, Lichao and others},
  booktitle={The Thirteenth International Conference on Learning Representations},
  year={2024}
}

@article{satbench,
  title = {SATBench: Benchmarking LLMs' Logical Reasoning via Automated Puzzle Generation from SAT Formulas},
  author = {Wei, Anjiang and Wu, Yuheng and Wan, Yingjia and Suresh, Tarun and Tan, Huanmi and Zhou, Zhanke and Koyejo, Sanmi and Wang, Ke and Aiken, Alex},
  journal = {arXiv preprint arXiv:2505.14615},
  year = {2025}
}

@article{math500,
  title = {Measuring Mathematical Problem Solving With the MATH Dataset},
  author = {Hendrycks, Dan and Burns, Collin and Kadavath, Saurav and Arora, Akul and Basart, Steven and Tang, Eric and Song, Dawn and Steinhardt, Jacob},
  journal = {arXiv preprint arXiv:2103.03874},
  year = {2021}
}

@article{bbeh,
  title = {BIG‑Bench Extra Hard},
  author = {Kazemi, Mehran and Fatemi, Bahare and Bansal, Hritik and Palowitch, John and Anastasiou, Chrysovalantis and Mehta, Sanket and Jain, Lalit D. and Aglietti, Virginia and Jindal, Disha and Chen, Peter and Dikkala, Nishanth and Tyen, Gladys and Liu, Xin and Shalit, Uri S. and Tay, Yi and Tran, Vinh Q. and Le, Quoc V. and Firat, Orhan},
  journal = {arXiv preprint arXiv:2502.19187},
  year = {2025}
}

@misc{amc2023,
  title = {AMC 2023 benchmark (American Mathematics Competition 2023)},
  author = {MAA and Benchmark users},
  year = {2023},
  note = {Used as evaluation benchmark in LLM reasoning research}
}

@misc{aime2024,
  title = {AIME 2024 benchmark (American Invitational Mathematics Examination 2024)},
  author = {MAA and Benchmark users},
  year = {2024},
  note = {Used as evaluation benchmark in LLM reasoning research}
}

@misc{aime2025,
  title = {AIME 2025 benchmark (American Invitational Mathematics Examination 2025)},
  author = {MAA and Benchmark users},
  year = {2025},
  note = {Used in recent benchmarks like SATBench analysis, etc.}
}

@article{olympiadbench,
  title={Olympiadbench: A challenging benchmark for promoting agi with olympiad-level bilingual multimodal scientific problems},
  author={He, Chaoqun and Luo, Renjie and Bai, Yuzhuo and Hu, Shengding and Thai, Zhen Leng and Shen, Junhao and Hu, Jinyi and Han, Xu and Huang, Yujie and Zhang, Yuxiang and others},
  journal={arXiv preprint arXiv:2402.14008},
  year={2024}
}

@article{GRPO,
  title   = {DeepSeekMath: Pushing the Limits of Mathematical Reasoning in Open Language Models},
  author  = {Shao, Zhihong and Wang, Peiyi and Zhu, Qihao and Xu, Runxin and Song, Junxiao and Bi, Xiao and Zhang, Haowei and Zhang, Mingchuan and Li, Y. K. and Wu, Y. and Guo, Daya},
  journal = {arXiv preprint arXiv:2402.03300},
  year    = {2024},
  doi     = {10.48550/arXiv.2402.03300},
  url     = {https://arxiv.org/abs/2402.03300},
  note    = {Introduces Group Relative Policy Optimization (GRPO)}
}

@misc{chatgpt-4o,
  title        = {GPT-4o System Card},
  author       = {{OpenAI}},
  year         = {2024},
  howpublished = {\url{https://arxiv.org/abs/2410.21276}},
  note         = {Model referenced in paper as ChatGPT-4o},
}

@misc{chatgpt-o3,
  title        = {OpenAI o3 and o4-mini System Card},
  author       = {{OpenAI}},
  year         = {2025},
  month        = {April},
  howpublished = {\url{https://openai.com/index/o3-o4-mini-system-card/}},
  note         = {Model referenced in paper as ChatGPT-o3},
}

@misc{chatgpt-5,
  title        = {Introducing GPT-5},
  author       = {{OpenAI}},
  year         = {2025},
  month        = {August},
  howpublished = {\url{https://openai.com/index/introducing-gpt-5/}},
  note         = {Model referenced in paper as ChatGPT-5},
}

@misc{seed-16,
  title        = {Seed1.6 Tech Introduction},
  author       = {{ByteDance}},
  year         = {2025},
  month        = {June},
  howpublished = {\url{https://seed.bytedance.com/en/seed1_6}},
  note         = {Model referenced in paper as Seed1.6},
}

@techreport{gemini25,
  title       = {Gemini 2.5: Pushing the Frontier with Advanced Reasoning, Multimodality, Long Context, and Next Generation Agentic Capabilities},
  author      = {{Gemini Team}},
  year        = {2025},
  month       = {June},
  institution = {Google DeepMind},
  url         = {https://storage.googleapis.com/deepmind-media/gemini/gemini_v2_5_report.pdf},
  note        = {Covers Gemini 2.5 Pro (and 2.5 Flash); model referenced in paper as Gemini-2.5-pro},
}

@misc{gemini20,
  title        = {Gemini 2.0 Flash},
  author       = {{Gemini Team}},
  year         = {2025},
  howpublished = {\url{https://cloud.google.com/vertex-ai/generative-ai/docs/models/gemini/2-0-flash}},
  note         = {Model referenced in paper as Gemini-2.0-flash},
}

@misc{gemini3pro,
  title        = {Gemini 3 Pro: the frontier of vision AI},
  author       = {{Gemini Team}},
  year         = {2025},
  howpublished = {\url{https://blog.google/technology/developers/gemini-3-pro-vision/}},
  note         = {Model referenced in paper as Gemini-3-pro},
}

@misc{claude35,
  title        = {Introducing Claude 3.5 Sonnet},
  author       = {{Anthropic}},
  year         = {2024},
  month        = {June},
  howpublished = {\url{https://www.anthropic.com/news/claude-3-5-sonnet}},
  note         = {Announcement; see also model card addendum},
}

@misc{claude4,
  title        = {Claude Opus 4 \& Claude Sonnet 4: System Card},
  author       = {{Anthropic}},
  year         = {2025},
  howpublished = {\url{https://www.anthropic.com/claude-4-system-card}},
  note         = {Model referenced in paper as Claude-4-sonnet},
}

@misc{GLM_Z1,
  title        = {GLM-Z1-32B-0414},
  author       = {{THUDM}},
  year         = {2025},
  howpublished = {\url{https://huggingface.co/THUDM/GLM-Z1-32B-0414}},
  note         = {GLM-Z1 reasoning model series (incl. 9B/32B); model card.}
}

@article{Qwen25,
  title   = {Qwen2.5 Technical Report},
  author  = {An Yang and Baosong Yang and Beichen Zhang and Binyuan Hui and Bo Zheng and Bowen Yu and Chengyuan Li and Dayiheng Liu and Fei Huang and Haoran Wei and Huan Lin and Jian Yang and Jianhong Tu and Jianwei Zhang and Jianxin Yang and Jiaxi Yang and Jingren Zhou and Junyang Lin and Kai Dang and Keming Lu and Keqin Bao and Kexin Yang and Le Yu and Mei Li and Mingfeng Xue and Pei Zhang and Qin Zhu and Rui Men and Runji Lin and Tianhao Li and Tianyi Tang and Tingyu Xia and Xingzhang Ren and Xuancheng Ren and Yang Fan and Yang Su and Yichang Zhang and Yu Wan and Yuqiong Liu and Zeyu Cui and Zhenru Zhang and Zihan Qiu and the Qwen Team},
  journal = {arXiv preprint arXiv:2412.15115},
  year    = {2024},
  url     = {https://arxiv.org/abs/2412.15115}
}

@article{Qwen3,
  title   = {Qwen3 Technical Report},
  author  = {An Yang and Anfeng Li and Baosong Yang and Beichen Zhang and Binyuan Hui and Bo Zheng and Bowen Yu and Chang Gao and Chengen Huang and Chenxu Lv and Chujie Zheng and Dayiheng Liu and Fan Zhou and Fei Huang and Feng Hu and Hao Ge and Haoran Wei and Huan Lin and Jialong Tang and Jian Yang and Jianhong Tu and Jianwei Zhang and Jianxin Yang and Jiaxi Yang and Jing Zhou and Jingren Zhou and Junyang Lin and Kai Dang and Keqin Bao and Kexin Yang and Le Yu and Lianghao Deng and Mei Li and Mingfeng Xue and Mingze Li and Pei Zhang and Peng Wang and Qin Zhu and Rui Men and Ruize Gao and Shixuan Liu and Shuang Luo and Tianhao Li and Tianyi Tang and Wenbiao Yin and Xingzhang Ren and Xinyu Wang and Xinyu Zhang and Xuancheng Ren and Yang Fan and Yang Su and Yichang Zhang and Yinger Zhang and Yu Wan and Yuqiong Liu and Zekun Wang and Zeyu Cui and Zhenru Zhang and Zhipeng Zhou and Zihan Qiu and the Qwen Team},
  journal = {arXiv preprint arXiv:2505.09388},
  year    = {2025},
  url     = {https://arxiv.org/abs/2505.09388}
}

@article{DeepSeek_R1_2025,
  title   = {DeepSeek-R1: Incentivizing Reasoning Capability in LLMs via Reinforcement Learning},
  author  = {Daya Guo and Dejian Yang and Haowei Zhang and Junxiao Song and Ruoyu Zhang and Runxin Xu and Qihao Zhu and Shirong Ma and Peiyi Wang and Xiao Bi and Xiaokang Zhang and Xingkai Yu and Yu Wu and Zongfang Wu and Zhibin Gou and Zhihong Shao and Zhuoshu Li and Ziyi Gao and {DeepSeek-AI Team} and others},
  journal = {arXiv preprint arXiv:2501.12948},
  year    = {2025},
  url     = {https://arxiv.org/abs/2501.12948},
  note    = {Covers DeepSeek-R1/-R1-Zero and distilled variants.}
}

@misc{liu2025synlogicsynthesizingverifiablereasoning,
      title={SynLogic: Synthesizing Verifiable Reasoning Data at Scale for Learning Logical Reasoning and Beyond}, 
      author={Junteng Liu and Yuanxiang Fan and Zhuo Jiang and Han Ding and Yongyi Hu and Chi Zhang and Yiqi Shi and Shitong Weng and Aili Chen and Shiqi Chen and Yunan Huang and Mozhi Zhang and Pengyu Zhao and Junjie Yan and Junxian He},
      year={2025},
      eprint={2505.19641},
      archivePrefix={arXiv},
      primaryClass={cs.AI},
      url={https://arxiv.org/abs/2505.19641}, 
}

\appendix

\label{sec:appendix}

\section{Supplement to Dataset Construction} \label{sec:dataset}

% This supplement provides additional details about (1) the prompt templates used for seed puzzle selection (Figure~\ref{fig:prompt_seed})  and downstream experiments (Figure~\ref{fig:prompt_wrapper}), and (2) the heuristic approach to measuring the puzzle difficulty (Section \ref{sec:difficulty}), as referenced in Section~3.1 and 3.4 of the main paper.
% Note that each template is presented in both its original Chinese form (used in practice) and its English translation (for readability).
This supplement provides additional details on the heuristic puzzle difficulty metric
and its validation, as well as the prompt templates used in this work.
% for seed puzzle selection (Figure~\ref{fig:prompt_seed})  and downstream experiments (Figure~\ref{fig:prompt_wrapper}).

\subsection{Puzzle Difficulty Metric} \label{sec:difficulty}

As described in Section~\ref{sec:diff_assess}, the difficulty metric is computed from four features: \texttt{sym\_num}, \texttt{cond\_num}, \texttt{desc\_len}, and \texttt{vars\_scale}.
Each variable $v$ is associated with a \texttt{diff\_factor} (Figure~\ref{fig:pipeline}B),
which specifies the direction of correlation between the variable value and puzzle difficulty:
a larger value makes the puzzle harder ($+1$), easier ($-1$), or has no effect ($0$, by default).
Algorithm~\ref{alg:difficulty} details the procedure for computing the final difficulty score.
% For a variable $v$ with domain $[v_{\min}, v_{\max}]$, its min-max normalized value is given by $\hat{v} = (v - v_{\min}) / ({v_{\max} - v_{\min}})$. The difficulty-adjusted value, $v_{adj}$, is subsequently defined as $\hat{v}$ or $1-\hat{v}$ 
% for positive or negative $diff\_factor$ values, respectively.
% % Note that variables whose corresponding $diff\_factor$ is $0$ will be excluded. 
% The $\texttt{vars\_scale}$ is computed as the average of all $v_{adj}$ for variables where $diff\_factor \neq 0$. 
% The final difficulty score is the mean of the min-max normalized (to $[0,1]$) values of 
% the four indicators.
% $sym\_num$, $cond\_num$, $desc\_len$, and $vars\_scale$. 

\begin{figure}[b]
  \centering
  \includegraphics[width=\linewidth]{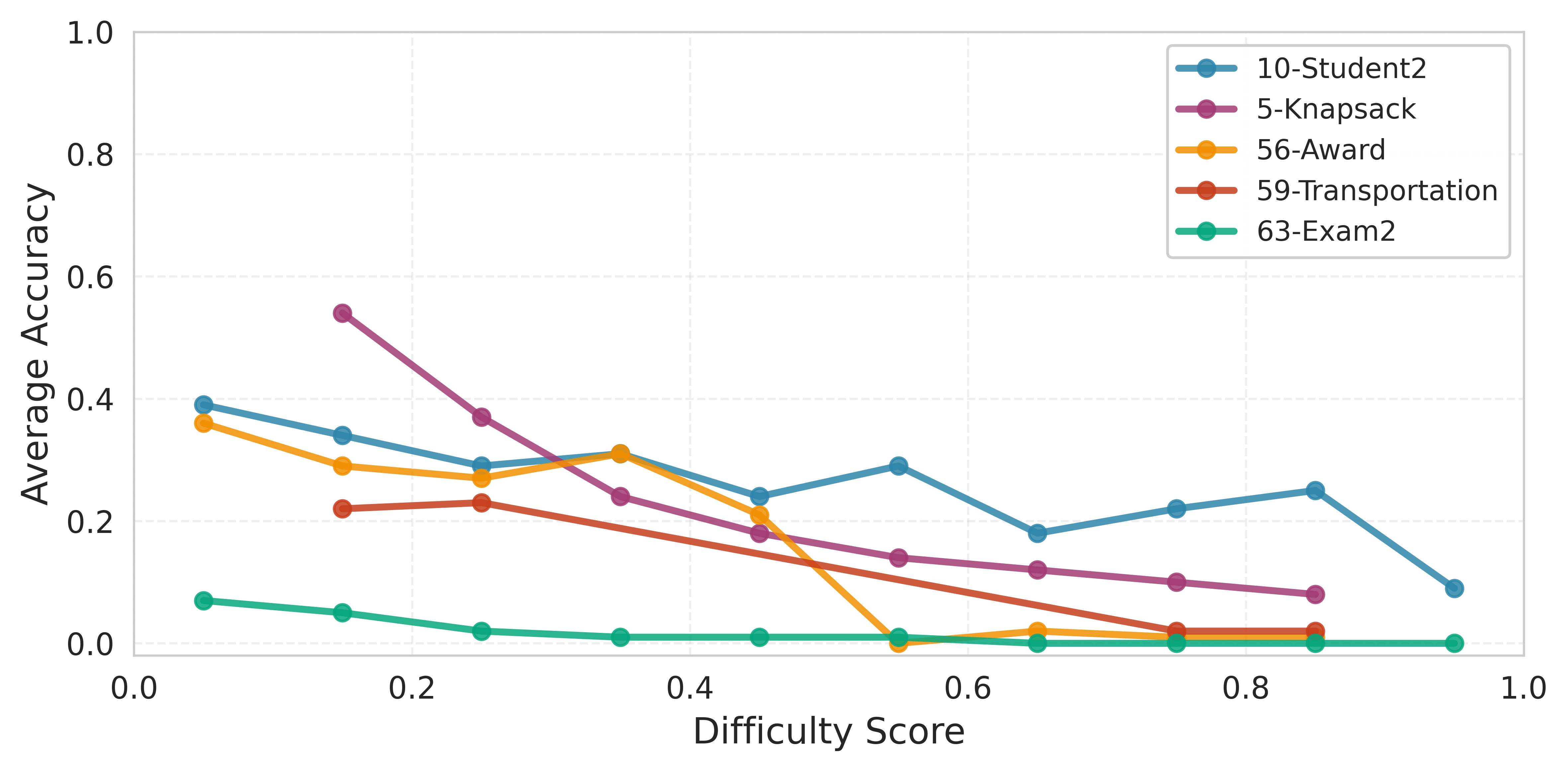}
  \caption{Average performance of four models on puzzles synthesized from 5 sampled seeds. The mid point is used as the difficulty score for each bin. Missing points indicate that no puzzles exist in the correponding bin.}
  \label{fig:diff_metric}
\end{figure}

To evaluate the alignment between our proposed metric and empirical difficulty, we conducted a study using four models from the Qwen2.5 family (7B, 14B, 32B, and 72B Instruct). We tested these models on puzzles synthesized from five randomly selected seeds in PuzzleClone, each seed leading to between 947 and 1,000 unique puzzles after deduplication. To mitigate the impact of randomness and ensure objective results, we conducted 10 independent trials for each model on every puzzle instance. Puzzles were then categorized into bins with a 0.1 difficulty score increment, and the mean accuracy was calculated for each group. As illustrated in Figure \ref{fig:diff_metric}, our metric demonstrates a strong inverse relationship with model performance. Specifically, the Pearson correlation coefficients range from -0.816 to -0.980, confirming that the metric accurately reflects task difficulty.

\begin{algorithm}[t]
\caption{Compute Puzzle Difficulty Score}
\label{alg:difficulty}
\begin{algorithmic}[1]
\REQUIRE Variables $\mathcal{V}$ from the puzzle \texttt{config}, features $sym\_num$, $cond\_num$, $desc\_len$, together with their minimum and maximum values computed over all puzzle instances
\ENSURE Difficulty score $D \in [0,1]$

\STATE Define $\textsc{MinMaxNorm}(x) = \dfrac{x - x_{\min}}{x_{\max} - x_{\min}}$

\STATE $\mathcal{A} \leftarrow \emptyset$ \COMMENT{Adjusted variable values}

\FOR{each variable $v \in \mathcal{V}$}
    \IF{$diff\_factor(v) = 0$}
        \STATE \textbf{continue}
    \ENDIF
    \STATE $\hat{v} \leftarrow \textsc{MinMaxNorm}(v)$
    \IF{$diff\_factor(v) > 0$}
        \STATE $v_{adj} \leftarrow \hat{v}$
    \ELSE
        \STATE $v_{adj} \leftarrow 1 - \hat{v}$
    \ENDIF
    \STATE add $v_{adj}$ to $\mathcal{A}$
\ENDFOR

\STATE $vars\_scale \leftarrow \text{mean}(\mathcal{A})$

\STATE $\hat{s} \leftarrow \textsc{MinMaxNorm}(sym\_num)$
\STATE $\hat{c} \leftarrow \textsc{MinMaxNorm}(cond\_num)$
\STATE $\hat{d} \leftarrow \textsc{MinMaxNorm}(desc\_len)$
\STATE $\hat{v} \leftarrow \textsc{MinMaxNorm}(vars\_scale)$

\STATE $D \leftarrow \text{mean}(\hat{s}, \hat{c}, \hat{d}, \hat{v})$
\RETURN $D$
\end{algorithmic}
\end{algorithm}

\subsection{Prompt Templates} \label{sec:prompts}
\begin{figure*}[b]
  \centering
  \includegraphics[width=\linewidth]{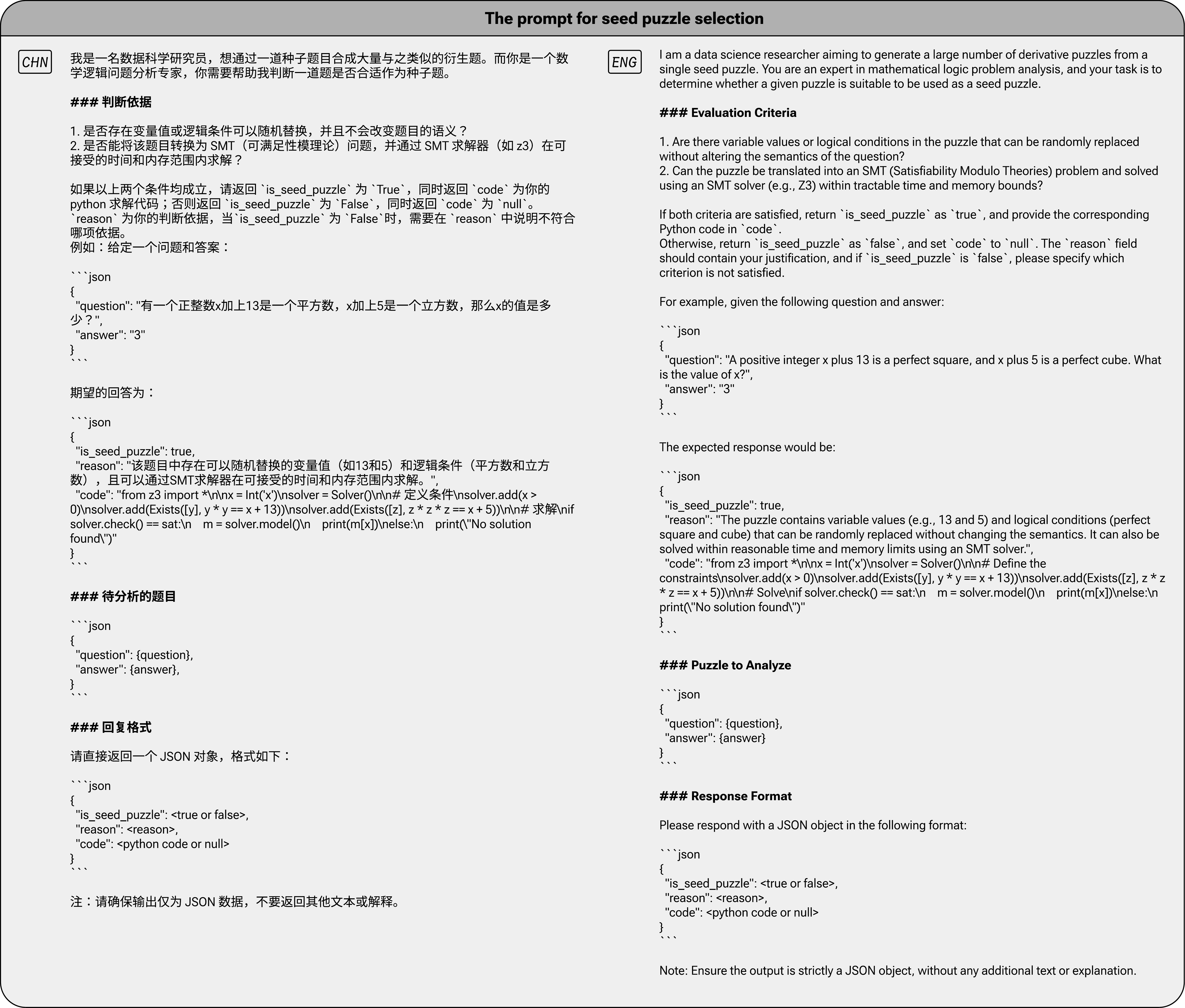}
  \caption{Seed Puzzle Selection Prompt for filtering seed puzzles via the Qwen2.5-72B-Instruct model.}
  \label{fig:prompt_seed}
\end{figure*}

\begin{figure}[t]
  \centering
  \includegraphics[width=\linewidth]{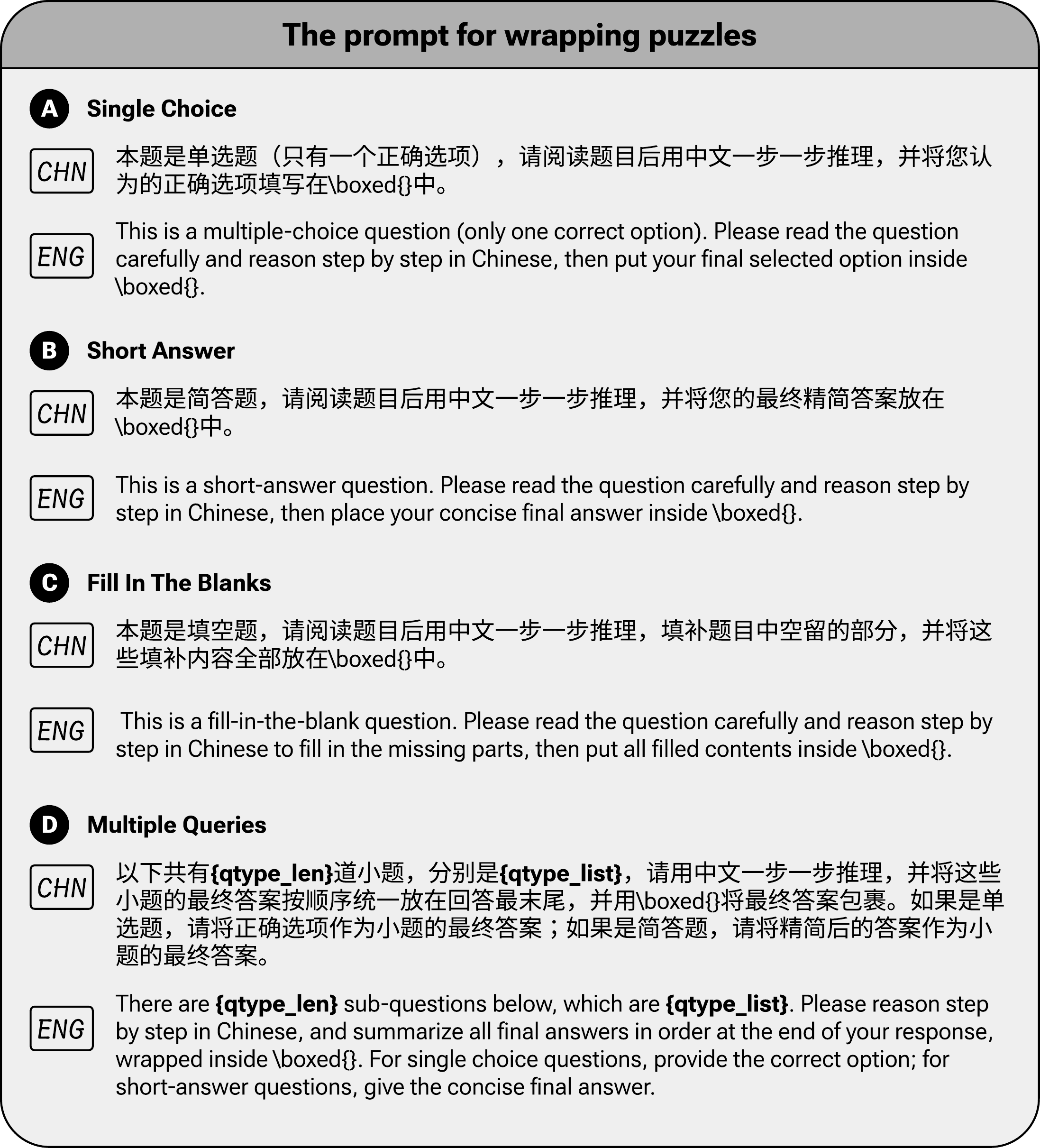}
  \caption{Question-Type Prompt Wrappers for LLM reasoning/training.}
  \label{fig:prompt_wrapper}
\end{figure}

Figure~\ref{fig:prompt_seed} and Figure~\ref{fig:prompt_wrapper} show the prompt templates used for seed puzzle selection and downstream experiments, respectively. 
Each template is presented in both its original Chinese form (used in practice) and its English translation (for readability).

\begin{figure*}[b]
  \centering
  \includegraphics[width=\linewidth]{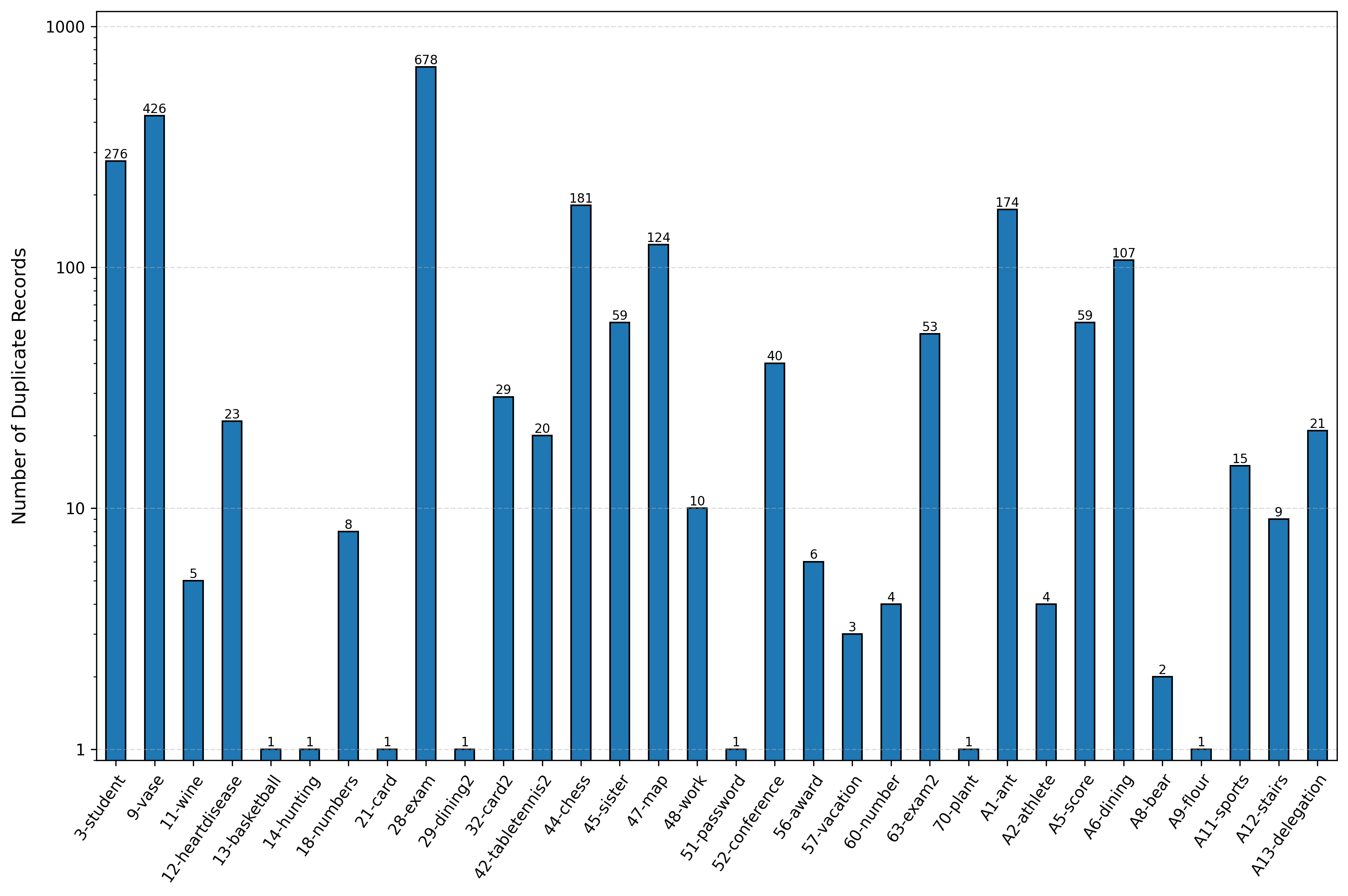}
  \caption{Distribution of duplicated instances across 32 seed puzzles. Only seed puzzles with at least one duplicate are shown.}
  \label{fig:duplication}
\end{figure*}

\section{Benchmark Statistics} \label{sec:data_stats}

\subsection{Question Type Distribution}
\label{sec:question_type_dist}

Our benchmark encompasses three primary question formats: short-answer, multiple-choice, and fill-in-the-blank questions. Table~\ref{tab:question_type_dist} presents the distribution of these question types across different data splits. When a puzzle contains multiple sub-questions, each sub-question is counted independently according to its type. 

The benchmark demonstrates a predominance of short-answer questions (75,486 instances), which require models to perform complex reasoning and generate precise responses. Multiple-choice questions constitute the second-largest category (18,929 instances), providing evaluation through option selection. Fill-in-the-blank questions, while less common (2,000 instances), test the model's ability to complete partial puzzle descriptions. This distribution reflects the natural complexity spectrum of logic puzzles, where open-ended reasoning challenges are most prevalent.

\begin{table*}[t]
\centering
\begin{tabular}{@{} l c c c c @{}}
\toprule
\textbf{Data Split} & \textbf{Short Answer} & \textbf{Multiple-Choice} & \textbf{Fill-in-the-Blank} & \textbf{Total} \\
\midrule
\texttt{normal/RL\_validate.jsonl} & 390 & 95 & 10 & 495 \\
\texttt{normal/Test.jsonl} & 5,108 & 1,262 & 124 & 6,494 \\
\texttt{normal/SFT.jsonl}\textsuperscript{*} & 1,966 & 475 & 50 & 2,491 \\
\texttt{normal/RL\_train.jsonl} & 45,195 & 11,202 & 1,099 & 57,496 \\
\midrule
\texttt{hard/RL\_validate.jsonl} & 390 & 95 & 10 & 495 \\
\texttt{hard/Test.jsonl} & 2,514 & 644 & 78 & 3,236 \\
\texttt{hard/SFT.jsonl}\textsuperscript{*} & 1,934 & 475 & 50 & 2,459 \\
\texttt{hard/RL\_train.jsonl} & 21,889 & 5,631 & 679 & 28,199 \\
\midrule
\textbf{Total}\textsuperscript{\textdagger} & \textbf{75,486} & \textbf{18,929} & \textbf{2,000} & \textbf{96,415} \\
\bottomrule
\end{tabular}
\caption{Distribution of question types across data splits in PuzzleClone. \textsuperscript{*}SFT samples are subsets selected from other training splits. \textsuperscript{\textdagger}Total excludes SFT splits to avoid double-counting.}
\label{tab:question_type_dist}
\end{table*}

\subsection{Answer Type Distribution}
\label{sec:answer_type_dist}

Beyond question format diversity, our benchmark exhibits rich structural variation in answer types. Table~\ref{tab:answer_type_dist} categorizes answers into eight distinct types, reflecting the diverse reasoning outputs required by logic puzzles:

\begin{itemize}
    \item \textbf{numeral}: Numeric values or lists thereof (e.g., \textit{42}, \textit{[3, 7, 12]}).
    \item \textbf{option}: Single-letter option identifiers (e.g., \textit{A}, \textit{D}).
    \item \textbf{ordered array}: Sequences where element order is semantically meaningful (e.g., rankings, positions).
    \item \textbf{nominal}: Categorical labels or entity names (e.g., person names, object types).
    \item \textbf{unordered array}: Sets where element order is irrelevant (e.g., collections of items).
    \item \textbf{ooa\_numeral}: Two-dimensional numeric arrays with order preservation in both dimensions.
    \item \textbf{ooa\_nominal}: Two-dimensional nominal arrays with order preservation in both dimensions.
    \item \textbf{oua\_nominal}: Two-dimensional nominal arrays where only outer-dimension order matters.
\end{itemize}

This taxonomy enables fine-grained evaluation of model capabilities across different reasoning output structures. The distribution shows that simpler answer types (\textit{numeral} and \textit{option}) dominate, while complex two-dimensional structures (\textit{ooa\_*}, \textit{oua\_*}) represent challenging edge cases with 1,000 instances each, deliberately included to test advanced compositional reasoning.

\begin{table*}[t]
\centering
\small
\begin{tabular}{@{} l c c c c c c c c c @{}}
\toprule
\textbf{Data Split} & \textbf{num.} & \textbf{opt.} & \textbf{ord.} & \textbf{nom.} & \textbf{unord.} & \textbf{oon.} & \textbf{oonl.} & \textbf{oun.} & \textbf{Total} \\
\midrule
\texttt{normal/RL\_validate} & 180 & 95 & 80 & 70 & 55 & 5 & 5 & 5 & 495 \\
\texttt{normal/Test} & 2,089 & 1,262 & 1,232 & 961 & 727 & 55 & 77 & 91 & 6,494 \\
\texttt{normal/SFT}\textsuperscript{*} & 910 & 475 & 404 & 352 & 275 & 25 & 25 & 25 & 2,491 \\
\texttt{normal/RL\_train} & 18,444 & 11,202 & 10,936 & 8,503 & 6,442 & 483 & 681 & 805 & 57,496 \\
\midrule
\texttt{hard/RL\_validate} & 180 & 95 & 80 & 70 & 55 & 5 & 5 & 5 & 495 \\
\texttt{hard/Test} & 1,501 & 644 & 351 & 433 & 227 & 46 & 24 & 10 & 3,236 \\
\texttt{hard/SFT}\textsuperscript{*} & 890 & 475 & 396 & 348 & 275 & 25 & 25 & 25 & 2,459 \\
\texttt{hard/RL\_train} & 13,175 & 5,631 & 3,011 & 3,756 & 1,928 & 406 & 208 & 84 & 28,199 \\
\midrule
\textbf{Total}\textsuperscript{\textdagger} & \textbf{35,569} & \textbf{18,929} & \textbf{15,690} & \textbf{13,793} & \textbf{9,434} & \textbf{1,000} & \textbf{1,000} & \textbf{1,000} & \textbf{96,415} \\
\bottomrule
\end{tabular}
\caption{Distribution of answer types across data splits. Column headers are abbreviated: \textbf{num.}=numeral, \textbf{opt.}=option, \textbf{ord.}=ordered array, \textbf{nom.}=nominal, \textbf{unord.}=unordered array, \textbf{oon.}=ooa\_numeral, \textbf{oonl.}=ooa\_nominal, \textbf{oun.}=oua\_nominal. \textsuperscript{*}SFT samples are subsets from other splits. \textsuperscript{\textdagger}Total excludes SFT splits.}
\label{tab:answer_type_dist}
\end{table*}

\subsection{Duplicate Instance Distribution} \label{sec:duplicate_distribution}
The distribution of duplicate puzzle instances grouped by seed puzzles during the deduplication step (Section~\ref{sec:deduplication}) is shown in Figure~\ref{fig:duplication}.
Among the 86 seed puzzles, 32 had at least one duplicated variant. 
We observe that puzzles with limited randomizable space tend to exhibit higher duplication rates.
For example, \texttt{28-exam} exhibited the highest duplication count, with 678 repeated instances.

\section{DSL Schema} \label{sec:schema}
This appendix provides a detailed specification of the Domain-Specific Language (DSL) used for puzzle encoding and generation. The DSL is structured around a main \texttt{PuzzleTemplate} object, which integrates all necessary components for defining a puzzle's logic, parameters, and text. The following tables detail each of the core components.

% \subsubsection{PuzzleTemplate: The Root Object}
\subsection{PuzzleTemplate}
The root object, \texttt{PuzzleTemplate}, is the top-level container that orchestrates all other elements of the puzzle definition. Its fields are detailed in Table~\ref{tab:puzzletemplate}.

\begin{table*}[t]
\centering
\begin{tabularx}{\textwidth}{@{} L{3.5cm} l X @{}}
\toprule
\textbf{Field} & \textbf{Type} & \textbf{Description} \\
\midrule
\texttt{custom\_operator} & \texttt{dict} (opt) & A dictionary of custom operators. The key is the operator name (e.g., \texttt{"double"}), and the value is a string containing either a Python lambda function (e.g., \texttt{"lambda x: x*2"}) or the path to a Python file defining the operator. \\
\addlinespace
\texttt{variables} & \texttt{dict} & A dictionary defining the puzzle's parameters. The key is the variable name and the value is a \texttt{Variable} object. See Table~\ref{tab:variable} for details. \\
\addlinespace
\texttt{symbols} & \texttt{dict} (opt) & A dictionary defining the symbols to be solved. Keys are symbol group names. Values can be \texttt{DefinedSymbol}, \texttt{DerivedSymbol}, or \texttt{DerivedSymbols} objects. See Tables~\ref{tab:definedsymbol} and \ref{tab:derivedsymbol}. \\
\addlinespace
\texttt{conditions} & \texttt{dict} (opt) & A dictionary of constraints. Keys are condition names. Values can be \texttt{StaticCondition} or \texttt{DynamicCondition} objects. See Tables~\ref{tab:staticcondition} and \ref{tab:dynamiccondition}. \\
\addlinespace
\texttt{calc\_solution} & \texttt{bool} & If true (default), the solver will compute the set of valid solutions for the generated puzzle instance. \\
\addlinespace
\texttt{max\_solution} & \texttt{int} & The maximum number of valid symbol configurations to find. If the solver exceeds this limit (default 6000), it halts. This refers to satisfying all constraints, not the final answer to a query. \\
\addlinespace
\texttt{post\_generation} & \texttt{PostGen} (opt) & Defines operations to be performed after the initial solution is found, such as deriving new variables from the solution. See Table~\ref{tab:postgen}. \\
\addlinespace
\texttt{optimize} & \texttt{Optimize} (opt) & Defines an optimization objective for problems that require minimizing or maximizing a certain value. See Table~\ref{tab:optimize}. \\
\addlinespace
\texttt{queries} & \texttt{dict} (opt) & A dictionary of questions for the puzzle. The key is the query name and the value is a \texttt{Query} or a selection-based query object. See Tables~\ref{tab:query} and \ref{tab:queryselection}. \\
\addlinespace
\texttt{desc} & \texttt{str} & A natural language template for the puzzle's introductory text, which can include placeholders for variables. \\
\bottomrule
\end{tabularx}
\caption{Schema for the \texttt{PuzzleTemplate} Root Object.}
\label{tab:puzzletemplate}
\end{table*}

\subsection{Variables}
\label{sec:dsl_variables}
The \texttt{Variable} object defines the parameters of a puzzle. A variable can be defined either by its type and domain or by a formula, as shown in Table~\ref{tab:variable}.

\begin{table*}[t]
\centering
\begin{tabularx}{\textwidth}{@{} L{3.5cm} l X @{}}
\toprule
\textbf{Field} & \textbf{Type} & \textbf{Description} \\
\midrule
\texttt{type} & \texttt{str} (opt) & The data type of the variable, e.g., \texttt{"int"}, \texttt{"bool"}. Must be defined if \texttt{formula} is not defined. \\
\addlinespace
\texttt{domain} & \texttt{str} (opt) & A string representing the variable's value space. Examples: \texttt{"[1, 10]"} for integers, \texttt{"['red', 'blue']"} for string options. Must be defined if \texttt{formula} is not defined. \\
\addlinespace
\texttt{formula} & \texttt{str} (opt) & A Python expression string to compute the variable's value. Example: \texttt{"randint(1,6) + randint(1,6)"}. If defined, \texttt{type} and \texttt{domain} must be omitted. \\
\bottomrule
\end{tabularx}
\caption{Schema for the \texttt{Variable} Object.}
\label{tab:variable}
\end{table*}

\subsection{Symbols}
\label{sec:dsl_symbols}
Symbols represent the quantities to be solved. They can be directly defined (\texttt{DefinedSymbol}) or derived from existing variables (\texttt{DerivedSymbol}).

\begin{table*}[t]
\centering
\begin{tabularx}{\textwidth}{@{} L{3.5cm} l X @{}}
\toprule
\textbf{Field} & \textbf{Type} & \textbf{Description} \\
\midrule
\texttt{source} & \texttt{list[str]} & A list of string expressions that serve as primary keys. For example, if \texttt{source} is \texttt{["children"]} and the variable \texttt{children} is \texttt{["Alice", "Bob"]}, two symbols are created. \\
\addlinespace
\texttt{attr} & \texttt{list[str]} (opt) & A list of attribute names for the symbol (e.g., \texttt{["color", "size"]}). If not provided, the symbol is a simple value. If provided, the symbol is a dictionary-like object with these attributes. \\
\addlinespace
\texttt{type} & \texttt{str} or \texttt{list[str]} & The Z3 type(s) for the symbol (e.g., \texttt{"Int"}, \texttt{"Bool"}). Must be a single string if \texttt{attr} is not defined. Must be a list of strings matching the length of \texttt{attr} if it is defined. \\
\addlinespace
\texttt{desc} & \texttt{str} or \texttt{list[str]} (opt) & A natural language description template. Follows the same format constraints as \texttt{type} based on the presence of \texttt{attr}. \\
\bottomrule
\end{tabularx}
\caption{Schema for the \texttt{DefinedSymbol} Object.}
\label{tab:definedsymbol}
\end{table*}

\begin{table*}[t]
\centering
\begin{tabularx}{\textwidth}{@{} L{3.5cm} l X @{}}
\toprule
\textbf{Field} & \textbf{Type} & \textbf{Description} \\
\midrule
\texttt{source} & \texttt{list[str]} & Data sources for selection. Each string is an expression evaluating to a list (e.g., a variable name or a literal list like \texttt{"[1, 2, 3]"}). \\
\addlinespace
\texttt{amount} & \texttt{list[str]} (opt) & The number of items to select from each corresponding source. If omitted, one item is selected from each source. Length must match \texttt{source}. \\
\addlinespace
\texttt{order} & \texttt{list[bool]} (opt) & If true for a source, selection order matters (permutation). If false, it does not (combination). Defaults to all true. Length must match \texttt{source}. \\
\addlinespace
\texttt{duplicate} & \texttt{list[bool]} (opt) & If true for a source, items can be selected more than once. Defaults to all false. Length must match \texttt{source}. \\
\addlinespace
\texttt{domain} & \texttt{str} (opt) & The total number of symbols (selections) to generate. Can be an integer literal or a variable name. \\
\addlinespace
\texttt{domain\_cond} & \texttt{bool} & If true (default), identical symbol combinations are disallowed across the entire selection process. \\
\addlinespace
\texttt{dim} & \texttt{int} & The number of dimensions for the symbol (default 1). Useful for creating matrices of related symbols. \\
\addlinespace
\texttt{dim\_cond} & \texttt{list} (opt) & A list of lists specifying inter-dimensional constraints. Each inner list contains source indices whose selected values cannot be identical. \\
\addlinespace
\texttt{custom\_cond} & \texttt{list} (opt) & A list of custom constraint dictionaries. Each dictionary specifies a \texttt{scope} (\texttt{"domain"} or \texttt{"dim"}), a list of source \texttt{fields}, and a Python lambda \texttt{constraint} string. \\
\addlinespace
\texttt{formula} & \texttt{str} (opt) & A Python expression string that defines the Z3 constraint for the generated symbol. \\
\addlinespace
\texttt{desc} & \texttt{str} & A natural language description template for the set of generated symbols. \\
\bottomrule
\end{tabularx}
\caption{Schema for the \texttt{DerivedSymbol} Object.}
\label{tab:derivedsymbol}
\end{table*}

\subsection{Conditions}
\label{sec:dsl_conditions}
Conditions define the logical constraints of the puzzle, as either static (\texttt{StaticCondition}) or dynamically generated (\texttt{DynamicCondition}) rules.

\begin{table*}[t]
\centering
\begin{tabularx}{\textwidth}{@{} L{3.5cm} l X @{}}
\toprule
\textbf{Field} & \textbf{Type} & \textbf{Description} \\
\midrule
\texttt{formula} & \texttt{str} & The constraint logic as a Python expression string (e.g., \texttt{"x + y < 10"}). \\
\addlinespace
\texttt{desc} & \texttt{str} (opt) & The corresponding natural language description for the puzzle text. \\
\bottomrule
\end{tabularx}
\caption{Schema for the \texttt{StaticCondition} Object.}
\label{tab:staticcondition}
\end{table*}

\begin{table*}[t]
\centering
\begin{tabularx}{\textwidth}{@{} L{3.5cm} l X @{}}
\toprule
\textbf{Field} & \textbf{Type} & \textbf{Description} \\
\midrule
\texttt{source} & \texttt{list[str]} & Data sources for parameter selection, same as in \texttt{DerivedSymbol}. \\
\addlinespace
\texttt{amount} & \texttt{list[str]} (opt) & Number of items to select from each source, same as in \texttt{DerivedSymbol}. \\
\addlinespace
\texttt{domain} & \texttt{str} (opt) & The number of conditions to generate, specified as a range string (e.g., \texttt{"[1, 5]"}). If omitted, one condition is generated. \\
\addlinespace
\texttt{domain\_cond} & \texttt{bool} & If true (default), identical parameter combinations are disallowed. \\
\addlinespace
\texttt{custom\_cond} & \texttt{list} (opt) & Custom constraints on parameter selection, same as in \texttt{DerivedSymbol}. \\
\bottomrule
\end{tabularx}
\caption{Additional fields for the \texttt{DynamicCondition} Object.}
\label{tab:dynamiccondition}
\end{table*}

\subsection{Queries}
\label{sec:dsl_queries}
Queries define the questions posed to the user, supporting open-ended and multiple-choice formats.

\begin{table*}[t]
\centering
\begin{tabularx}{\textwidth}{@{} L{3.5cm} l X @{}}
\toprule
\textbf{Field} & \textbf{Type} & \textbf{Description} \\
\midrule
\texttt{desc} & \texttt{str} & The natural language text of the question. \\
\addlinespace
\texttt{ans\_formula} & \texttt{str} & A Python expression to compute the correct answer from the solution. \\
\addlinespace
\texttt{ans\_text} & \texttt{str} & A template for formatting the display of the answer. \\
\addlinespace
\texttt{ans\_assertion} & \texttt{str} (opt) & A Python expression that must evaluate to true for the answer to be considered valid (e.g., to ensure a unique solution exists). \\
\bottomrule
\end{tabularx}
\caption{Schema for the \texttt{Query} (Open-Ended) Object.}
\label{tab:query}
\end{table*}

\begin{table*}[t]
\centering
\begin{tabularx}{\textwidth}{@{} L{3.5cm} l X @{}}
\toprule
\textbf{Field} & \textbf{Type} & \textbf{Description} \\
\midrule
\multicolumn{3}{@{}l}{\textit{Base fields for all selection queries include \texttt{desc}, \texttt{query\_type}, \texttt{select\_type}, and \texttt{opt\_num}.}} \\ \addlinespace
\texttt{source} & \texttt{list[str]} & Data sources for generating option parameters. \\
\addlinespace
\texttt{cond} & \texttt{str} & Condition scope for an option to be correct/incorrect. \texttt{"any"}: satisfies at least one solution; \texttt{"all"}: satisfies all solutions. \\
\addlinespace
\texttt{opt\_formula} & \texttt{str} & A Python expression that evaluates the correctness of a generated option. \\
\addlinespace
\texttt{opt\_text} & \texttt{str} (opt) & A template for the display text of the option. \\
\addlinespace
\texttt{custom\_cond} & \texttt{list} (opt) & Custom constraints on option parameter selection. \\
\bottomrule
\end{tabularx}
\caption{Schema for the \texttt{QuerySelectionTemplate} Object (for multiple-choice options).}
\label{tab:queryselection}
\end{table*}

\subsection{Auxiliary Definitions}
\label{sec:dsl_auxiliary}
The DSL includes objects for post-processing and optimization tasks.

In addition, PuzzleClone supports several internal operators and reserved words for expressiveness. For instance, \texttt{\_opt} and \texttt{\_sym} refer to the parameter randomization result of each option and symbol, while \texttt{get\_faker} is an internal operator for fake entity generation.

\begin{table*}[ht]
\centering
\begin{tabularx}{\textwidth}{@{} L{3.5cm} l X @{}}
\toprule
\textbf{Field} & \textbf{Type} & \textbf{Description} \\
\midrule
\texttt{post\_gen\_vars} & \texttt{dict} (opt) & A dictionary to define new variables. Keys are new variable names, values are Python expressions to extract values from the solution. \\
\addlinespace
\texttt{post\_gen\_conditions} & \texttt{dict} (opt) & A dictionary to add new constraints. Keys are new constraint names, values are \texttt{StaticCondition} objects. \\
\bottomrule
\end{tabularx}
\caption{Schema for the \texttt{PostGen} Object.}
\label{tab:postgen}
\end{table*}

\begin{table*}[t]
\centering
\begin{tabularx}{\textwidth}{@{} L{3.5cm} l X @{}}
\toprule
\textbf{Field} & \textbf{Type} & \textbf{Description} \\
\midrule
\texttt{type} & \texttt{str} & The optimization type: \texttt{"minimize"} or \texttt{"maximize"}. \\
\addlinespace
\texttt{formula} & \texttt{str} & The formula representing the value to be optimized. \\
\bottomrule
\end{tabularx}
\caption{Schema for the \texttt{Optimize} Object.}
\label{tab:optimize}
\end{table*}

\section{Puzzle Specification Examples}
\label{sec:appendix_specs}

This section provides the complete DSL specification files for three representative puzzles from our benchmark. All descriptive texts, originally in Chinese, have been translated to English.

\begin{compactitem}
    \item \texttt{1-hamburger} (Figure \ref{fig:pipeline}): See Listings \ref{lst:hamburger_part1} and \ref{lst:hamburger_part2}.

    \item \texttt{2-graduation}: See Listing \ref{lst:graduation}.

    \item \texttt{9-vase}: See Listing \ref{lst:vase}.

    \item \texttt{11-wine}: See Listing \ref{lst:wine}.

    \item \texttt{23-product}: See Listing \ref{lst:product}.

    \item \texttt{28-exam} See Listing \ref{lst:exam}.
\end{compactitem}

\begin{listing}[t]
\begin{lstlisting}[ caption={DSL Specification for the ``Hamburger'' Puzzle (Part 1/2).}, label={lst:hamburger_part1}, basicstyle={\scriptsize\ttfamily}]
variables:
  s_num: {type: int, domain: "[4, 7]", diff_factor: 1}
  f_num: {type: int, domain: "[3, 5]", diff_factor: 1}
  names: {formula: get_faker(s_num, 'name')}
  food: {formula: get_faker(f_num, 'food')}
symbols:
  buy: {source: [names, food], type: bool}
conditions:
  purchased_at_least_one_kind:
    formula: And([Or([buy[(p, f)] for f in food]) for p in names])
    desc: "{s_num} students, {', '.join(names)}, have purchased at least one kind of food: {','.join(food)}."
  if_a_then_not_b:
    source: [food, "[False, True]"]
    amount: ['2', '2']
    formula: >
      And([Implies(buy[(p, _sym[0][0])] if _sym[1][0] else Not(buy[(p, _sym[0][0])]), buy[(p, _sym[0][1])] if _sym[1][1] else Not(buy[(p, _sym[0][1])])) for p in names])
    desc: >
      People who {'did not buy' if _sym[1][0] else 'bought'} {_sym[0][0]} {'did not buy' if _sym[1][1] else 'bought'} {_sym[0][1]}.
  at_least_one_person_bought:
    source: [food]
    domain: "[1, 2]"
    formula: Or([buy[(p, _sym[0])] for p in names])
    desc: At least one person bought {_sym[0]}.
  a_bought_b:
    source: [names, food, "[False, True]"]
    domain: "[s_num*f_num//3, s_num*f_num//2]"
    formula: buy[(_sym[0], _sym[1])] == _sym[2]
    desc: "{_sym[0]} {'bought' if _sym[2] else 'did not buy'} {_sym[1]}."
  a_b_exclusive:
    source: [names]
    amount: ['2']
    domain: "[1, 2]"
    formula: >
      And([Implies(buy[(_sym[0][0], f)], Not(buy[(_sym[0][1], f)])) for f in food])
    desc: "{_sym[0][1]} did not buy any items that {_sym[0][0]} bought."
  assumption:
    source: [names, "range(2, f_num)"]
    amount: ['2', '1']
    formula: And([Sum([If(buy[(p, f)], 1, 0) for f in food]) == _sym[1][0] for p in
      _sym[0]])
    desc: "If both {_sym[0][0]} and {_sym[0][1]} bought {_sym[1][0]} kinds of products,"
\end{lstlisting}
\end{listing}

\begin{listing}[tb]
\begin{lstlisting}[caption={DSL Specification for the ``Hamburger'' Puzzle (Part 2/2).}, label={lst:hamburger_part2}, basicstyle={\scriptsize\ttfamily}]
# --- (Continuation) Spec of "1-hamburger" ---
queries:
  question:
    desc: "which of the following must be true?"
    opt_num: 4
    templates:
    - source: [names]
      amount: ['2']
      cond: all
      opt_formula: >
        sum([1 for f in food if get_value(_model, And(buy[(_opt[0][0], f)], buy[(_opt[0][1], f)]))]) == 1
      opt_text: >
        There is exactly one food item that both {_opt[0][0]} and {_opt[0][1]} bought.
    - source: [names, food, "[False, True]"]
      amount: ['1', '1', '1']
      cond: all
      opt_formula: get_value(_model, buy[(_opt[0][0], _opt[1][0])]) == _opt[2][0]
      opt_text: "{_opt[0][0]} {'bought' if _opt[2][0] else 'did not buy'} {_opt[1][0]}."
desc: >
  {purchased_at_least_one_kind}Their choices satisfy these conditions: {if_a_then_not_b}{at_least_one_person_bought}{a_bought_b}{a_b_exclusive}{assumption}{question}
\end{lstlisting}
\end{listing}

\begin{listing}[tb]
\begin{lstlisting}[caption={DSL specification for the ``Graduation'' puzzle.}, label={lst:graduation}, basicstyle={\scriptsize\ttfamily}]
variables:
  p_num: {type: int, domain: "[6, 12]", diff_factor: 1}
  select_num: {type: int, domain: "[p_num // 2 - 1, p_num // 2 + 1]"}
  names: {formula: generate_letters(p_num)}
  name_desc: {formula: "', '.join(names)"}
symbols:
  events: # Represents whether a person is selected
    source: [names]
    type: bool
    desc: "{_names} was selected for the ceremony"
conditions:
  base: # Total number of selected people
    formula: gen_event_count_condition(events, 'equal', select_num)
    desc: "Select {select_num} people for the graduation ceremony."
  cond1: # XOR condition
    source: [events]
    domain: "[1, 3]"
    amount: ['2']
    formula: gen_event_count_condition(_sym[0], 'equal', 1)
    desc: "Either {get_p(_sym[0][0], 'names')} or {get_p(_sym[0][1], 'names')} must be selected, but not both."
  cond2: # Implication condition
    source: [events]
    domain: "[1, 3]"
    amount: ['2']
    formula: Implies(_sym[0][1], _sym[0][0])
    desc: "Unless {get_p(_sym[0][0], 'names')} is selected, {get_p(_sym[0][1], 'names')} cannot be."
queries:
  question:
    source: [events]
    desc: "Which of the following could be a valid selection?"
    opt_num: 5
    amount: [select_num]
    cond: any
    opt_formula: sum([get_value(_model, _opt[0][i]) for i in range(select_num)]) == select_num
    opt_text: "{', '.join(get_p(_opt[0], 'names'))}"
  q2:
    source: [events]
    desc: "The selected group must include:"
    opt_num: 4
    amount: ['2']
    cond: all
    opt_formula: sum([get_value(_model, _opt[0][i]) for i in range(2)]) >= 1
    opt_text: "{get_p(_sym[0][0], 'names')} or {get_p(_sym[0][1], 'names')}."
  q3:
    source: [events]
    desc: "Which two people cannot be selected at the same time?"
    opt_num: 5
    amount: ['2']
    cond: all
    opt_formula: sum([get_value(_model, _opt[0][i]) for i in range(2)]) <= 1
    opt_text: "{get_p(_sym[0][0], 'names')} and {get_p(_sym[0][1], 'names')}"
desc: "From {p_num} graduates ({name_desc}), {base}. The selection must satisfy these conditions: {cond1} {cond2} {queries}"
\end{lstlisting}
\end{listing}

\begin{listing}[t]
\begin{lstlisting}[caption={DSL specification for the ``Vase'' puzzle.}, label={lst:vase}, basicstyle={\scriptsize\ttfamily}]
variables:
  p_num: {type: int, domain: "[3, 10]", diff_factor: 1}
  broken_vase_num: {type: int, domain: "[1, round(p_num / 3)]"}
  names: {formula: get_faker(p_num, 'name')}
  name_desc: {formula: "', '.join(names)"}
symbols:
  names_s: # Who broke the vase
    source: [names]
    type: bool
    desc: "{_names} broke the vase"
  speeches_s: # What each person said
    source: [names_s, "[True, False]", "['eq']"]
    domain: p_num
    domain_cond: false # Allow duplicate speeches
    formula: make_expr(_sym[2], _sym[0], _sym[1])
    desc: >
      {names[_index]} says: "{get_p(_sym[0], 'names')} 
      {'broke' if _sym[1] else 'did not break'} the vase"
conditions:
  cond1: # Total number of culprits
    formula: gen_event_count_condition(names_s, 'equal', broken_vase_num)
    desc: >
      The mother knows {broken_vase_num} of {p_num} children broke the vase.
  cond2: # Culprits are liars
    formula: "[Implies(names_s[names[i]], Not(speeches_s[i])) for i in range(p_num)]"
    desc: "The children who broke the vase are definitely lying."
queries:
  question:
    desc: "Who broke the vase?"
    ans_formula: >
      get_p(get_TF_events_for_each_solution(names_s, _solutions, True), 'names')
    ans_text: "','.join(_ans[0])"
    ans_assertion: len(_ans) == 1
desc: >
  There are {p_num} children: {name_desc}. {broken_vase_num} broke a vase. 
  Their statements: {', '.join(get_desc(speeches_s))}. {cond1}, and {cond2}. {question}
\end{lstlisting}
\end{listing}

\begin{listing}[tb]
\begin{lstlisting}[caption={DSL specification for the ``Wine'' puzzle.}, label={lst:wine}, basicstyle={\scriptsize\ttfamily}]
variables:
  wine_num: {type: int, domain: "[6, 12]", diff_factor: 1}
  beer_num: {type: int, domain: "[1, 2]"}
  bought_wine_of_first_customer: {type: int, domain: "[1, wine_num/2]"}
  vol_times: {type: int, domain: "[2, 5]"}
  wines: {formula: generate_letters(wine_num)}
symbols:
  wine_s: # Each barrel has a volume and belonging (0:beer, 1:cust1, 2:cust2)
    source: [wines]
    attr: [volume, belonging]
    type: [int, int]
conditions:
  wine_belonging: {formula: "And([Or(x==0, x==1, x==2) for x in wine_s.get('belonging')])"}
  wine_0:
    formula: "Sum([If(x==0, 1, 0) for x in wine_s.get('belonging')]) == beer_num"
    desc: "{beer_num} barrels contain beer."
  wine_1:
    formula: "Sum([If(x==1, 1, 0) for x in wine_s.get('belonging')]) == bought_wine_of_first_customer"
    desc: "The first customer bought {bought_wine_of_first_customer} barrels of wine."
  wine_times:
    formula: >
      Sum([If(x1 == 2,x2,0) for x1,x2 in zip(wine_s.get('belonging'), wine_s.get('volume'))])
      == vol_times * Sum([If(x1==1,x2,0) for x1,x2 in zip(wine_s.get('belonging'), wine_s.get('volume'))])
    desc: "The second customer bought {vol_times} times the volume of the first."
  wine_volume_domain: {formula: "And([And(x > 0, x <= 50) for x in wine_s.get('volume')])"}
  wine_volume_distinct: {formula: "gen_event_count_condition(wine_s.get('volume'), 'distinct')"}
post_generation: # Solve for volumes first, then fix them
  post_gen_vars:
    vol: "get_value(_sol, wine_s.get('volume'))"
  post_gen_conditions:
    vol_cond:
      formula: "And([wine_s[w].get('volume') == vol[i] for i, w in enumerate(wines)])"
queries:
  question:
    source: [range(0, wine_num)]
    desc: "Which barrel(s) contain beer? Please provide the letters for the correct options."
    opt_num: 4
    amount: [beer_num]
    cond: all
    opt_formula: "sum([get_value(_model, wine_s[wines[_opt[0][i]]].get('belonging')) == 0 for i in range(beer_num)]) == beer_num"
    opt_text: "{','.join([str(vol[_opt[0][i]]) for i in range(beer_num)])}"
desc: >
  A merchant has {wine_num} barrels of wine and beer with volumes: {','.join([str(v) + ' gallons' for v in vol])}.
  There are {wine_num - beer_num} barrels of wine. {wine_0}, {wine_1}, and {wine_times}.
  No wine is left. {question}
\end{lstlisting}
\end{listing}

\begin{listing}[tb]
\begin{lstlisting}[caption={DSL specification for the ``Product'' puzzle.}, label={lst:product}, basicstyle={\scriptsize\ttfamily}]
variables:
  p_num: {type: int, domain: "[6, 17]", diff_factor: 1}
  products: {formula: get_faker(p_num, 'product')}
  names: {formula: get_faker(1, 'name')}
symbols:
  pos: # The position of each product on a conveyor belt
    source: [products]
    type: int
conditions:
  pos_domain: {formula: "And([And(pos[x] >= 1, pos[x] <= p_num) for x in products])"}
  pos_distinct: {formula: "gen_event_count_condition(pos, 'distinct')"}
  cond1: # Relative distance between products
    source: [products, "range(1, p_num - 2)"]
    amount: ['2', '1']
    domain: "[p_num // 2, p_num]"
    formula: "Or(pos[_sym[0][0]] - pos[_sym[0][1]] == _sym[1][0], pos[_sym[0][1]] - pos[_sym[0][0]] == _sym[1][0])"
    desc: "{_sym[0][0]} and {_sym[0][1]} have {'no' if _sym[1][0] == 1 else str(_sym[1][0] - 1)} items between them."
  cond2: # Adjacent products
    source: [products]
    amount: ['2']
    domain: "[1, p_num // 2]"
    formula: "pos[_sym[0][1]] - pos[_sym[0][0]] == 1"
    desc: "{_sym[0][1]} is placed immediately after {_sym[0][0]}."
  cond3: # Absolute position constraint
    source: [products, "range(1, p_num+1)"]
    amount: ['1', '1']
    domain: "[0, p_num // 3]"
    formula: "pos[_sym[0][0]] != _sym[1][0]"
    desc: "{_sym[0][0]} is not in position {_sym[1][0]}."
max_solution: 600
queries:
  question:
    source: [products, "range(1, p_num+1)"]
    desc: "If the statements are true, which of the following must be true?"
    opt_num: 6
    amount: ['1', '1']
    cond: all
    opt_formula: "get_value(_model, pos[_opt[0][0]]) == _opt[1][0]"
    opt_text: "{_opt[0][0]} is in position {_opt[1][0]}."
desc: >
  After shopping, {names[0]} placed {p_num} items on a conveyor belt, 
  ordered from front to back (pos 1 to {p_num}). {conditions} {question}
\end{lstlisting}
\end{listing}

\begin{listing}[ht]
\begin{lstlisting}[caption={DSL specification for the ``Exam'' puzzle.}, label={lst:exam}, basicstyle={\scriptsize\ttfamily}]
variables:
  p_num: {type: int, domain: "[4, 10]", diff_factor: 1}
  exam_num: {type: int, domain: "[2, p_num]", diff_factor: 1}
  num_performed_well: {type: int, domain: "[1, p_num - 1]"}
  names: {formula: get_faker(p_num, 'name')}
  exams: {formula: get_faker(exam_num, 'major')}
symbols:
  performed_well: {source: [names], type: bool}
  passed: {source: [names, exams], type: bool}
conditions:
  num_performed_well_cond:
    formula: gen_event_count_condition(performed_well, 'equal', num_performed_well)
    desc: "Only {num_performed_well} of {p_num} people performed well."
  cond1:
    source: [exams]
    amount: ['2']
    domain: "[p_num, p_num]"
    domain_cond: false
    formula: >
      And(Implies(performed_well[names[_index]], passed[(names[_index], _sym[0][0])]),
          Implies(Not(performed_well[names[_index]]), Not(passed[(names[_index], _sym[0][1])])))
    desc: '{names[_index]} says: "If I perform well, I will pass {_sym[0][0]}. If not, I will fail {_sym[0][1]}."'
  only_one_who_passed_some_exam:
    source: ["['False', 'True']"]
    amount: ['2']
    domain: "[2, 2]"
    formula: >
      Or([Sum([If(And(passed[(p, e)] == _sym[0][1], performed_well[p] != _sym[0][0]),
          1, 0) for p in names]) == 0 for e in exams])
    desc: >
      For one subject, only those who performed {'well' if _sym[0][0] else 'poorly'}
      {'' if _sym[0][1] else 'did not '}pass.
max_solution: 500
queries:
  question:
    desc: "Who performed well?"
    ans_formula: to_unique([[p for p in names if get_value(_model, performed_well[p])] for _model in _solutions])
    ans_text: "','.join(_ans[0])"
    ans_assertion: len(_ans) <= 1
desc: >
  {p_num} people ({', '.join(names)}) took {exam_num} exams ({', '.join(exams)}).
  {num_performed_well_cond} Statements: {cond1} Also: {only_one_who_passed_some_exam} {question}
\end{lstlisting}
\end{listing}

\section{Training Parameters} \label{sec:training_param}
Table \ref{tab:sft_hparams} outlines SFT setup: 32K context, cosine LR schedule with warmup, bf16/tf32, gradient checkpointing, ZeRO-3 Offload via DeepSpeed, six epochs, small per-device batch with accumulation, and step-based checkpointing.

\begin{table*}[t]
\centering
\small
\begin{tabular}{llp{7.2cm}}
\toprule
\textbf{Category} & \textbf{Setting} & \textbf{Value} \\
\midrule
Objective
 & Task & Supervised fine-tuning (prompt: \texttt{chat\_template}) \\
\midrule
Model \& Optimization
 & Base model & \texttt{Qwen2.5-7B-Instruct} \\
 & Model max length & 32{,}768 \\
 & Per-device batch size & 1 \\
 & Grad accumulation steps & 4 \\
 & Learning rate & $1\times10^{-5}$ \\
 & Weight decay & 0.05 \\
 & Warmup ratio & 0.03 \\
 & LR scheduler & cosine \\
\midrule
Regularization \& Checkpointing
 & Gradient checkpointing & True \\
 & Filter by length & True \\
 & Save only model & True \\
\midrule
Precision
 & Numeric formats & bf16=True; tf32=True \\
\midrule
Distributed / System
 & DeepSpeed & ZeRO-3 Offload \\
 & Dataloader workers & 1 \\
\midrule
Training \& Logging
 & Epochs & 6 \\
 & Evaluation & \texttt{eval\_strategy=no} \\
 & Saving & \texttt{save\_strategy=steps}; \texttt{save\_steps=615}; \texttt{save\_total\_limit=999} \\
 & Logging & \texttt{logging\_steps=1}; \texttt{report\_to=none} \\

\midrule
Compute
 & GPUs & 16 (8 H100 per node $\times$ 2) \\
\bottomrule
\end{tabular}
\caption{Key hyperparameters for SFT training.}
\label{tab:sft_hparams}
\end{table*}

Table \ref{tab:grpo_hparams} summarizes key GRPO hyperparameters: KL regularization, vLLM rollouts, fsdp2 distributed training, Qwen2.5-7B-Instruct, five epochs, sixteen H100 across two nodes, dynamic batching and token-limits.

\begin{table*}[ht]
\centering
\small
\begin{tabular}{llp{6.8cm}}
\toprule
\textbf{Category} & \textbf{Setting} & \textbf{Value} \\
\midrule
Algorithm
 & RL type & PPO with GRPO advantage estimator \\
 & KL regularization & use\_kl\_loss=True; kl\_loss\_coef=0.001; kl\_loss\_type=low\_var\_kl; use\_kl\_in\_reward=False; entropy\_coeff=0 \\
 & Critic warmup & 0 \\
\midrule
Model \& Optimization
 & Base model & \texttt{Qwen2.5-7B-Instruct} \\
 & Learning rate & $1\times10^{-6}$ \\
 & Gradient checkpointing & True \\
 & PPO mini-batch size & 128 \\
 & Dynamic batch size & True \\
 & Max token len / GPU (PPO) & 24{,}000 \\
\midrule
Rollout / Inference
 & Engine & vLLM; use\_remove\_padding=True \\
 & Generations per prompt & $n=5$ \\
 & Max batched tokens (rollout) & 5{,}120 \\
 & GPU memory utilization & 0.6 \\
 & Tensor model parallel size & 2 \\
\midrule
Distributed / FSDP
 & Strategy & fsdp2 (actor / ref / critic / reward model) \\
 & Offload & actor: param=False, optimizer=False; ref: param=True \\
\midrule
Training \& Logging
 & Epochs & 5 \\
 & Validation & val\_before\_train=True \\
 & Save / Test freq (steps) & save=290; test=80 \\
 & Logger & console, tensorboard \\
\midrule
Compute
 & GPUs & 16 (8 H100 per node $\times$ 2) \\
\bottomrule
\end{tabular}
\caption{Key hyperparameters for GRPO training.}
\label{tab:grpo_hparams}
\end{table*}

% \subsection{Case Study}

\begin{figure*}[b]
  \centering
  \includegraphics[width=\linewidth]{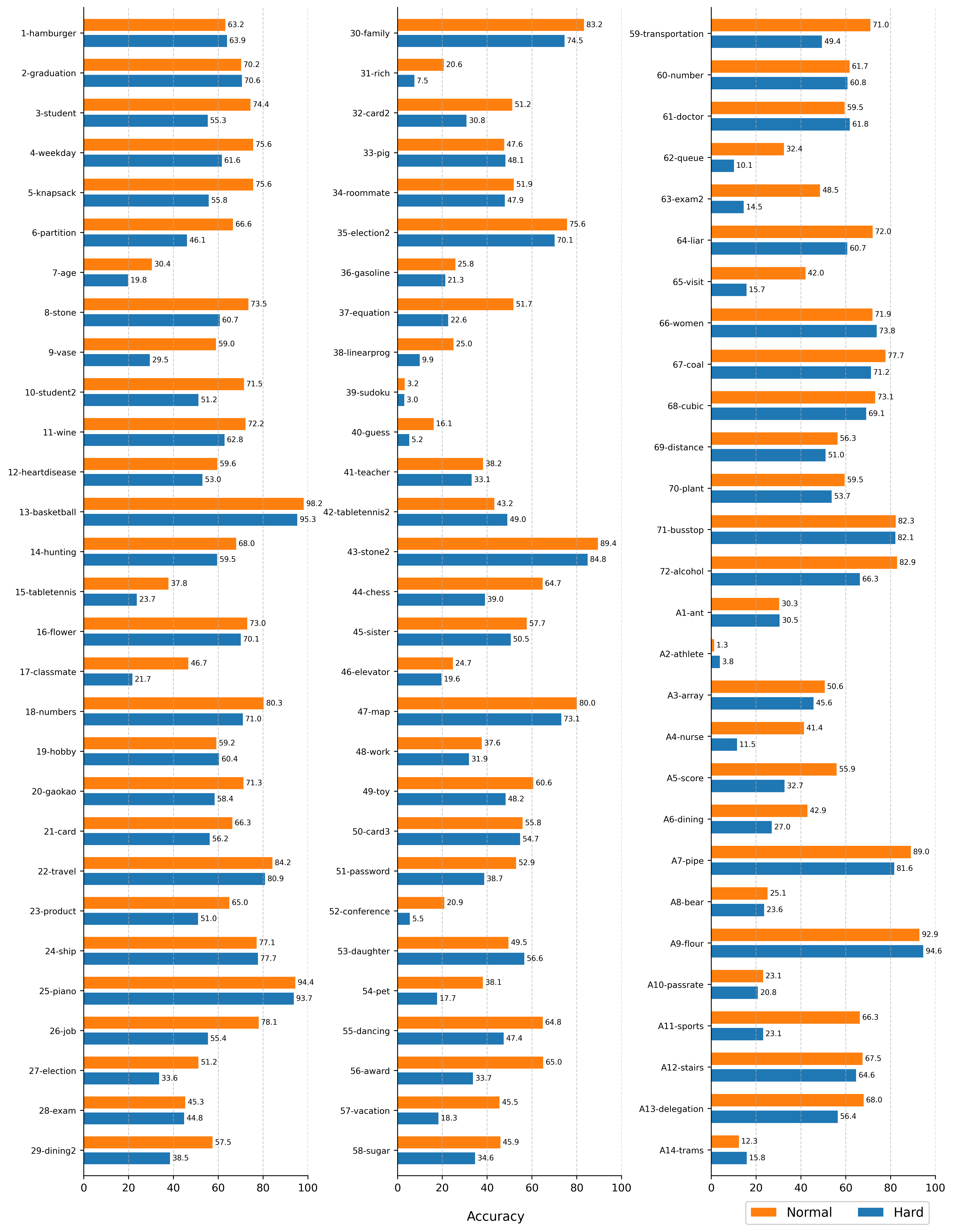}
  \caption{The average accuracy of all models on the PuzzleClone test set grouped by the seed puzzles.}
  \label{fig:accuracy_all}
\end{figure*}

\section{Supplement to Experimental Results}

Figure~\ref{fig:accuracy_all} presents the average accuracy of all evaluated models on the PuzzleClone test set, grouped by their originating seed puzzles.
% Unexpectedly, for certain seed puzzles (e.g., \texttt{2-graduation}), the variants in the \textit{hard} subset exhibit higher average accuracy than those in the \textit{normal} subset. This counterintuitive pattern primarily stems from limitations in our current difficulty scoring mechanism, which may not fully capture semantic or structural complexity. For a detailed discussion on this issue, please refer to the Revisiting Difficulty Estimation part in Section 6.
For most seed puzzles, the \textit{hard} variants yield lower average accuracy compared to the \textit{normal} ones, demonstrating the effectiveness of our difficulty stratification. However, for certain seed puzzles, the \textit{hard} subset shows unexpectedly higher accuracy. This counterintuitive trend reveals limitations in our current difficulty scoring method, which may not fully capture the underlying reasoning complexity. For a more detailed analysis, please refer to the Revisiting Difficulty Estimation part in Section 6.

\section{Details of Dynamic Rephrasing}
This section introduces \textit{dynamic rephrasing}, a key functionality of PuzzleClone that enables transforming the original puzzle into new languages or scenarios in a fully accurate and verifiable way. 

Given an original puzzle $P$ generated by a pair of specification and configuration files ($Q_s$ and $Q_c$), PuzzleClones enabling adapting $P$ to a new Puzzle $P'$ based on a new specification file $Q_s'$, where $Q_s'$ and $Q_s$ differ only in the descriptive texts. During generation, the values of all variables and parameters of symbols, conditions, and queries will be directly extracted from $Q_c$ instead of through randomization. This new puzzle can be validated in a similar manner to the reproduced seed puzzle, ensuring its validity.

It is important to note that in a few cases, the values of some variables or parameters still need to be randomized, such as the names of people mentioned in the puzzle randomized by Faker. To accommodate such cases, PuzzleClone enables \textit{partial config-based generation}, enabling users to specify variables for randomization through command-line options.

\end{document}